\documentclass{article}

\usepackage[numbers]{natbib}

 \usepackage[preprint]{neurips_2026}

\usepackage[utf8]{inputenc} % allow utf-8 input
\usepackage[T1]{fontenc}    % use 8-bit T1 fonts
\usepackage{hyperref}       % hyperlinks
\usepackage{url}            % simple URL typesetting
\usepackage{booktabs}       % professional-quality tables
\usepackage{amsfonts}       % blackboard math symbols
\usepackage{nicefrac}       % compact symbols for 1/2, etc.
\usepackage{microtype}      % microtypography
\usepackage{xcolor}         % colors

\usepackage{graphicx}
\usepackage{multirow}
\usepackage{subcaption}
\usepackage{enumitem}
\usepackage{wrapfig}

\title{Shift-Aware Identity-Guided Latent Refinement for Referring Audio--Visual Segmentation}

\author{
\textbf{Kun Li}$^{1}$
\quad
\textbf{Sami Sebastian Brandt}$^{2}$
\quad
\textbf{Michael Ying Yang}$^{3}$
\\
$^1$ University of Liverpool, UK
\quad
$^2$ IT University of Copenhagen, DK
\quad
$^3$ University of Bath, UK
}

\begin{document}

\maketitle

\begin{abstract}
Referring Audio--Visual Segmentation (Ref-AVS) is an emergent multimodal task requiring the precise identification and segmentation of a specific object based on a natural language expression grounded in both visual and auditory cues.
Unlike traditional AVS, which segments generic sound sources, or referring video object segmentation (Ref-VOS), which relies solely on visual--textual grounding, Ref-AVS necessitates fine-grained tri-modal interaction across vision, audio, and language.
However, existing methods often fail to maintain consistency when subjected to target shift, a phenomenon where the model erroneously transfers its focus between objects due to transient occlusions, acoustic fading, or identity ambiguities.
To address this, we propose a novel \textbf{S}hift-\textbf{A}ware \textbf{I}dentity-guided \textbf{L}atent (\textbf{SAIL}) refinement framework.
It introduces a dual-token strategy that decouples frame-level dynamics from a global semantic anchor to explicitly model target-shift cues.
We design a shift-aware refinement module that rectifies drift by aligning tokens with domain-specific evidence.
Finally, we leverage a decoder that integrates the refined embeddings with a continuous memory update to maintain spatio-temporal coherence and identity consistency throughout the video sequence.
Extensive experiments conducted on two Ref-AVS benchmarks demonstrate the effectiveness of our proposed method, significantly outperforming existing AVS, Ref-VOS, and Ref-AVS approaches.
The code will be released after publication.
\end{abstract}

\section{Introduction}
\label{sec:intro}
The integration of visual and auditory modalities is fundamental to how humans perceive, interpret, and navigate the multifaceted complexity of dynamic real-world environments.
Consequently, robust multimodal understanding has become a cornerstone for a broad spectrum of intelligent systems, ranging from event localization \cite{um2025soundloc} and target tracking \cite{kang2025tracking} to holistic scene understanding \cite{lin2013holistic}.
Building upon these foundations, Referring Audio--Visual Segmentation (Ref-AVS) \cite{wang2024refavs} has emerged as a granular tri-modal task that requires not only understanding the language query provided by users but also precisely grounding a specific object within the spatio-temporal domain.
While tasks such as audio--visual question answering (AVQA) focus on high-level linguistic reasoning \cite{li2022musicavqa, li2026qstar, li2025post}, Ref-AVS demands a dense, pixel-level mask of a target identified through the synergistic alignment of visual, auditory, and textual cues.

\begin{figure}[t]
  \centering
  % \vspace{-3mm}
  \begin{subfigure}[t]{0.9\linewidth}
    \includegraphics[width=1\linewidth]{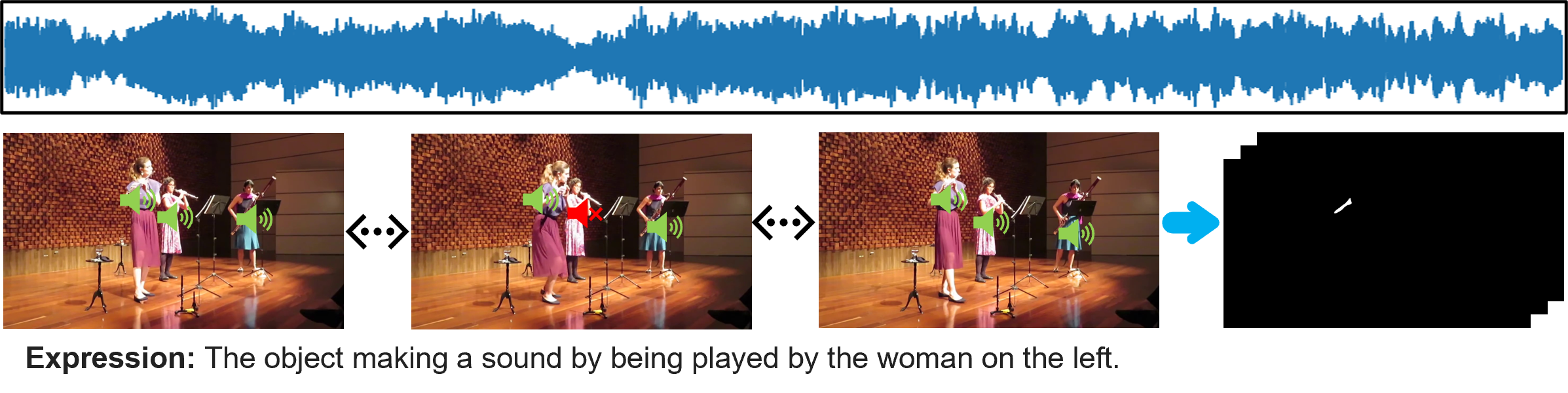}
    \subcaption{Referring Audio--Visual Segmentation (Ref-AVS) task.}
    \label{fig: refavs}
  \end{subfigure}
  % \hspace{2mm}
  \begin{subfigure}[t]{0.44\linewidth}
    \includegraphics[width=1\linewidth]{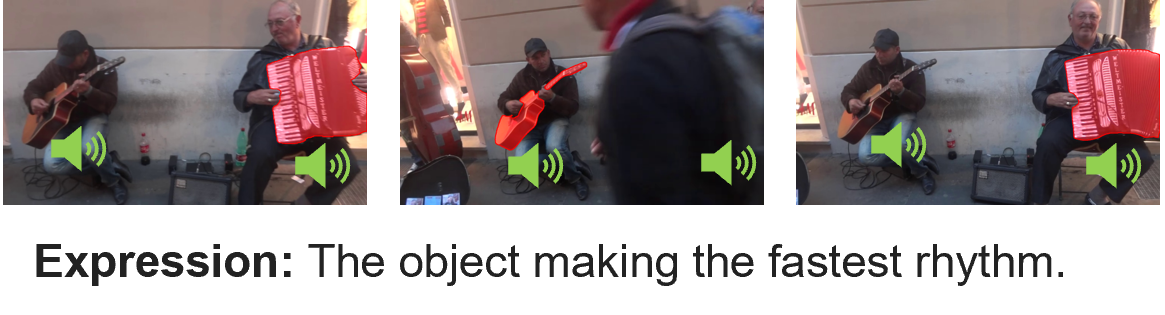}
    \subcaption{Target shift in the visual domain.}
    \label{fig: v-d}
  \end{subfigure}
  \hspace{1mm}
  \begin{subfigure}[t]{0.44\linewidth}
    \includegraphics[width=1\linewidth]{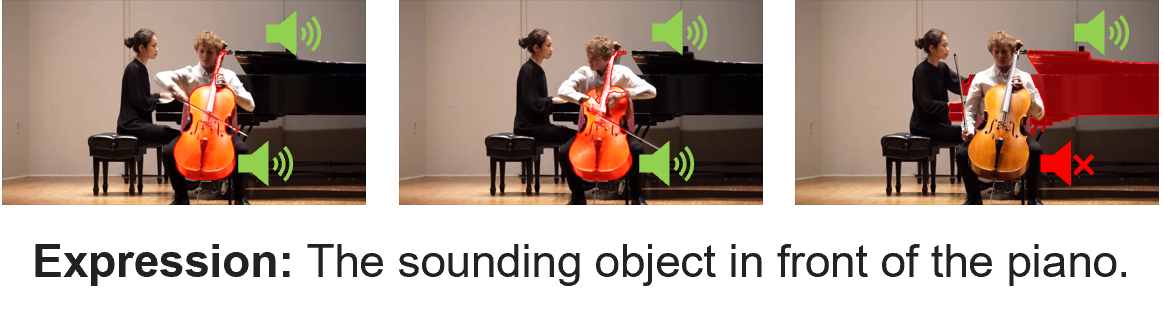}
    \subcaption{Target shift in the auditory domain.}
    \label{fig: a-d}
  \end{subfigure}
  % \hspace{2mm}
  \caption{Illustration of Ref-AVS and target shift problems.
    (a) Given a tri-modal input (audio, video, and expression), a Ref-AVS model performs sequential mask prediction for the referred object.
    (b) Visual target shift caused by transient occlusion or overshadowing (\textit{e.g.}, shifting from the accordion to a guitar).
    (c) Auditory target shift caused by intermittent sound fading or pauses (\textit{e.g.}, shifting from the cello to a piano).
    }
  \vspace{-5mm}
  \label{fig: illustration}
\end{figure}

Ref-AVS fundamentally differs from related multimodal segmentation tasks by requiring a dense, object-level grounding of \textit{tri-modal} signals to resolve perceptual ambiguities, as shown in Fig.~\ref{fig: refavs}.
While Audio--Visual Segmentation (AVS) \cite{gao2024avsegformer, zhou2022avs} focuses on segmenting generic sound sources at the category level, Ref-AVS demands the resolution of fine-grained object-level identities, particularly in complex polyphonic scenes with multiple active sounds.
Similarly, whereas Referring Video Object Segmentation (Ref-VOS) \cite{li2023r2vos, wu2022referformer} relies exclusively on visual--textual grounding, Ref-AVS necessitates tri-modal alignment to handle scenarios where visual cues alone are insufficient for target identification.
For example, in a scene featuring multiple violinists, only Ref-AVS can leverage a query like ``the violin playing the staccato notes'' to isolate the target by correlating specific auditory rhythms with localized visual motion.
This intersection of modalities makes Ref-AVS a higher-level task, testing a model's ability to fuse disparate sensory streams into a singular, cohesive object representation, thereby achieving a finer-grained level of video understanding.

Despite the conceptual strength of tri-modal grounding, a critical yet under-explored challenge in Ref-AVS is the \textit{target shift} problem.
We define target shift as the temporal drift of  predicted object identity throughout a video sequence.
Formally, target shift occurs when the segmentation mask at time $t$ corresponds to an object instance $O'$ that differs from the target $O$ referred to by the language query.
% , despite the presence of the correct target in the scene.
Such shifts typically arise when visual or auditory cues become temporarily unreliable (\textit{e.g.}, occlusion or sound absence), causing the model to re-anchor its prediction to another salient object.
% This phenomenon occurs when the model's focus drifts from the intended referent toward other modality-dominant objects, often triggered by transient conflicts among the audio, visual, and textual signals.
% In dynamic video sequences, the referred object frequently undergoes transient changes in its visibility, acoustic status, or contextual identity, leading to inconsistent temporal predictions.
For instance, as illustrated in Fig.~\ref{fig: v-d}, a target may be temporarily occluded or overshadowed by other entities;
in such cases, the model's focus may erroneously shift from the referred accordion to a dominant visual distractor, such as another sounding instrument (\textit{e.g.}, a guitar) that remains visually prominent.
% , or a person entering the same spatial location.
% , both of which can mislead the attention mechanism.
% to another sounding object that suddenly rushes into the center of the scene, becoming a dominant visual distractor.
% Similarly, the characteristic sound of a target may intermittently fade or pause (\textit{e.g.,} the cello and piano in Fig.~\ref{fig: a-d}), provoking a corresponding drift in the auditory dimension, especially when the auditory distractor explicitly presented in the textual guidance.
Similarly, a target's characteristic sound may intermittently fade or pause (\textit{e.g.}, the cello and piano in Fig.~\ref{fig: a-d}), triggering an auditory drift.
This is especially common when an auditory distractor is explicitly mentioned in the textual guidance, misleading the model's focus.
During these intervals of sensory ambiguity, maintaining a persistent identity is essential for a Ref-AVS model.

% Without a dedicated mechanism to anchor the target, 
Existing Ref-AVS models fail to handle the target shift problem.
% Existing Ref-AVS frameworks remain fundamentally ill-equipped to  due to a lack of long-term temporal reasoning.
% Traditional specialized multimodal approaches \cite{wang2024refavs, radman2025tsam} typically rely on localized spatial and temporal alignment, utilizing transformer-based \cite{vaswani2017attention} fusion to establish cross-modal correspondences.
Specialized multimodal approaches \cite{radman2025tsam, wang2024refavs} typically rely on localized spatial and temporal alignment to establish cross-modal correspondences.
While effective in relatively static scenes, these methods struggle to maintain object identity across sensory gaps, leading to fragmented masks or inconsistent tracking.
Recently, Multimodal Large Language Model (MLLM) based methods \cite{du2025crab, wang2025sam2love} attempt to achieve tri-modal alignment through implicit reasoning.
However, the generated global representation fails to capture fine-grained spatio-temporal details or consistently preserve the target during scene changes \cite{heo2023singletoken1, ying2023singletoken2}.
This absence of a robust, persistent target identity mechanism leaves a significant gap in Ref-AVS.
Furthermore, these MLLM approaches frequently depend on computationally expensive multi-stage training on additional datasets for instruction tuning \cite{luo2025aurora} or external object detectors \cite{zhou2025tgs}.

To address the limitations, we propose \textbf{SAIL}, a novel end-to-end MLLM-based framework that leverages anchor identity and shift refinement for the Ref-AVS task.
% Unlike prior MLLM-based approaches that rely on a single-token representation, SAIL leverages a dual-level tokenization scheme within the LLM to facilitate robust tri-modal alignment.
Specifically, we introduce a temporal \texttt{[seg]} token to capture frame-specific focus and a global \texttt{[query]} token to establish a persistent semantic anchor for the referred object.
By explicitly modeling the variance across visual and auditory modalities, our framework quantifies potential target drift through a dedicated shift-estimation mechanism.
Finally, these shift-refined embeddings are integrated with a weighted global anchor and then fed into a mask decoder.
By synchronizing these refined features with a continuous memory bank update, SAIL effectively mitigates target shift and ensures spatio-temporal identity consistency for mask prediction.
We validate our method on two benchmarks Ref-AVSBench \cite{wang2024refavs} and $\mathrm{R^2}$-AVSBench \cite{zhou2025tgs}, and the results show that SAIL significantly outperforms previous AVS, Ref-VOS, and Ref-AVS methods.
Our \textbf{main contributions} can be summarized as:
\begin{itemize}[leftmargin=*, itemsep=0pt]
\item We identify and formalize the \emph{target shift} problem in Ref-AVS, where transient cross-modal inconsistencies lead to unstable segmentation predictions across time.

\item We propose SAIL, a novel end-to-end MLLM-based framework for Ref-AVS that explicitly models object identity and mitigates modality-induced drift via shift-aware refinement.

\item We introduce an identity-guided token design with shift-aware update, which disentangles persistent anchor semantics from transient multimodal evidence and enforces consistent tri-modal grounding.

\item Extensive experiments demonstrate that SAIL achieves new state-of-the-art performance on the Ref-AVSBench and $\mathrm{R^2}$-AVSBench.
\end{itemize}

\section{Related work}
\label{sec: related work}

\noindent
\textbf{Referring Image Segmentation} (RIS) serves as a foundational task for language-driven pixel-level understanding, where the goal is to generate the binary mask of a target described by a text.
% Existing RIS methods \cite{ding2022vlt, li2025sbanet, liu2024rmsin} primarily focus on building effective cross-modal fusion modules to align linguistic and visual features.
Existing RIS methods \cite{ding2022vlt, li2025sbanet} primarily focus on effectively fusing visual and linguistic features.
For example, LAVT \cite{yang2022lavt} demonstrated the efficacy of early vision-language fusion within a transformer-based encoder.
% by deeply integrating linguistic cues into the visual backbone.
Referring Video Object Segmentation (Ref-VOS) extends this challenge into the temporal domain, requiring the model to localize the referred target while maintaining spatio-temporal consistency across multiple frames.
URVOS \cite{seo2020urvos} adopted a segment-then-track strategy, performing segmentation on key frames followed by mask propagation via cross-modal and memory attention mechanisms.
ReferFormer \cite{wu2022referformer} introduced language-conditioned object queries to track the referred objects across frames.
With the emergence of foundation models \cite{kirillov2023sam, radford2021clip}, recent approaches (\textit{e.g.,} LISA \cite{lai2024lisa} and VISA \cite{yan2024visa}) have leveraged MLLMs to achieve image- and video-level reasoning segmentation.
However, these methods focus solely on visual--textual grounding, which is insufficient for handling the complex and diverse dynamics of audio--visual scenes.

\noindent
\textbf{Audio--Visual Segmentation} (AVS) aims to segment the pixel-level regions of objects that are actively producing sound within a video sequence.
Unlike traditional video segmentation, AVS requires the model to establish a fine-grained correspondence between audio signal and spatial pixels \cite{zhou2025avs1, liu2024avs2}.
The task was formally pioneered by AVSBench \cite{zhou2022avs}, which introduced a benchmark for both single-source and multi-source sounding object segmentation.
AVSegFormer \cite{gao2024avsegformer} utilized a transformer-based backbone to treat the audio signal as a global query, attending to visual features to decode the object masks.
GAVS \cite{wang2024gavs} and SAMA \cite{liu2024sama} have explored the integration of Segment Anything Model (SAM) \cite{kirillov2023sam} to leverage its powerful zero-shot spatial priors, allowing for more precise boundary delineation of sounding entities.
% Other works have addressed the temporal consistency of sounding objects, ensuring that masks do not flicker during momentary audio pauses. These methods typically employ multi-modal fusion modules—such as cross-attention or gated fusion—to suppress non-sounding visual distractors.
By learning the visual representations of acoustic events, AVS provides a critical technical foundation for scene understanding, though it operates without the explicit high-level semantic guidance provided by natural language.

\noindent
\textbf{Referring Audio--Visual Segmentation} (Ref-AVS) is a recently emerged task requiring pixel-level identification of targets specified by the synergistic alignment of auditory, visual, and linguistic cues.
Wang \textit{et al.} \cite{wang2024refavs} established the foundational Ref-AVSBench dataset and proposed a baseline that adaptively combines tri-modal features for spatio-temporal correlation.
Subsequent research has sought to enhance mask fidelity by incorporating foundation models.
For instance, TSAM \cite{radman2025tsam} integrated SAM to exploit robust spatial and temporal priors, while SAM2-LOVE \cite{wang2025sam2love} leveraged SAM2 \cite{ravi2024sam2} by mapping multimodal representations into a learnable \texttt{[seg]} token for implicit reasoning, further enhanced through token propagation and accumulation.
These specialized multimodal methods prioritize the design of cross-modal fusion mechanisms to align the three modalities.
Recently, several works \cite{jin2025simtoken, luo2025aurora, ying2025oisa-1b} have shifted toward MLLMs to leverage their powerful reasoning capabilities.
Crab \cite{du2025crab} proposed a unified audio--visual scene understanding framework with explicit inter-task cooperation via interaction-aware LoRA \cite{hu2022lora} adapters.
TGS-Agent \cite{zhou2025tgs} introduced an explainable think-ground-segment pipeline; it employed LLMs to generate textual descriptions that guide object-level grounding via Grounding-DINO \cite{liu2024grounding}, followed by mask generation through SAM2.
However, existing methods often fail to account for the dynamic nature of audio--visual scenes where the referred object may undergo significant changes.
Differently, our method addresses the target shift problem by introducing a dual-token strategy and a shift-aware refinement mechanism, which anchors the referred object's identity across the timeline.

\begin{figure}[t]
    \centering
    \includegraphics[width=0.85\linewidth]{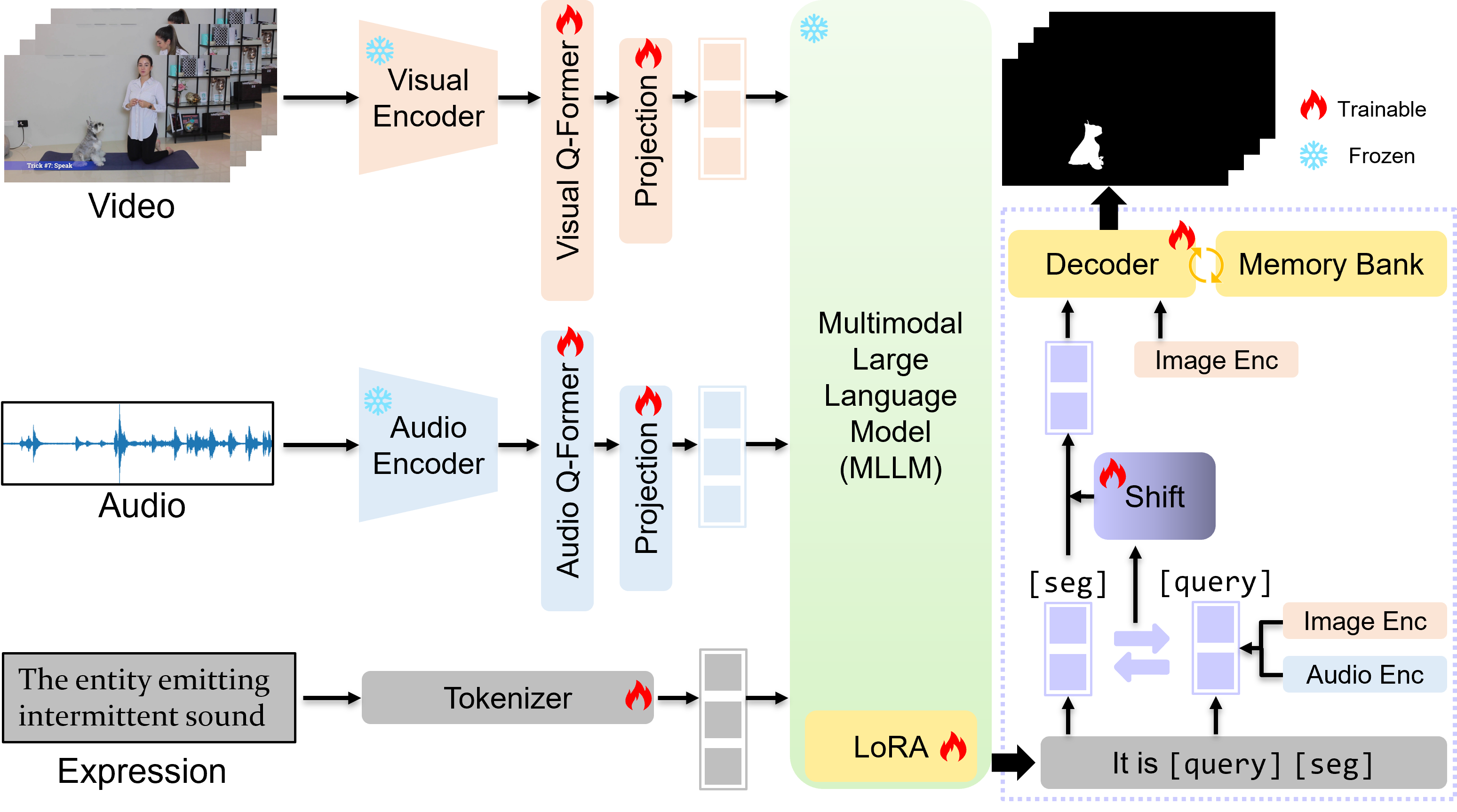}
    \caption{Overview of the proposed SAIL framework.
    % We employ a trainable Q-Former \cite{li2023blip2} and projection layers to tokenize features extracted from pre-trained visual and audio encoders (Sec.~\ref{sec: input representation}).
    % Our MLLM with LoRA \cite{hu2022lora} reasons over these unified tri-modal tokens using a dual-token strategy (Sec.~\ref{sec: dual tokens}), generating frame-level \texttt{[seg]} tokens for local dynamics and a global \texttt{[query]} token as a semantic anchor.
    % Domain-specific evidence is then leveraged to estimate the latent shift by comparing the current segmentation representation with the anchor's semantics.
    % The shift-aware refinement module (Sec.~\ref{sec: shift anchor}) rectifies this drift, ensuring the representation remains aligned with the referred anchor.
    % Finally, a SAM2-based \cite{ravi2024sam2} mask decoder (Sec.~\ref{sec: decoder}) utilizes the refined embeddings and a memory mechanism to generate consistent mask predictions for the entire video sequence.
    Our MLLM with LoRA \cite{hu2022lora} reasons over the unified tri-modal tokens, generating frame-level \texttt{[seg]} tokens for local dynamics and a global \texttt{[query]} token as a semantic anchor.
    The shift-aware refinement module estimates and rectifies the latent shift by comparing the domain-specific evidence with the anchor's semantics.
    The mask decoder utilizes the refined embeddings and a memory mechanism to generate consistent mask predictions for the entire video sequence.
    For brevity, projection units before the decoder are omitted.
    }
    \label{fig: framework}
    \vspace{-5mm}
\end{figure}

\section{Method}
\label{sec: method}

\subsection{Preliminary}
\label{sec: preliminary}
\noindent
\textbf{Task formulation.}
The task of Ref-AVS can be defined as follows.
Given a video sequence of $T$ frames $\mathcal{V} \in \mathbb{R}^{T \times 3 \times H \times W}$ and its synchronized audio track $\mathcal{A}$, along with a natural language query $\mathcal{Q}$, the objective is to learn a mapping function $\mathcal{F}_m: \{\mathcal{V}, \mathcal{A}, \mathcal{Q}\} \to M$ to generate a sequence of segmentation masks $M \in \mathbb{R}^{T \times H \times W}$ that identify the referent.
% , where $H$ and $W$ denote the height and width of each frame, respectively.
% Unlike standard referring tasks (\textit{e.g.}, RIS and Ref-VOS), Ref-AVS necessitates the dense interaction of tri-modal signals to obtain pixel-level predictions over dynamic video sequences.
Beyond grounding, the task requires the model to maintain identity permanence for the referred object across the entire video sequence.

\noindent
\textbf{Overall architecture.}
To address the challenges of Ref-AVS, we propose SAIL, a novel MLLM-based method with shift-aware refinement, thereby ensuring robust spatio-temporal identity consistency.
Fig.~\ref{fig: framework} illustrates the overall architecture of the proposed method, which consists of three main components: i) an MLLM incorporating intra-modality feature tokenization in Sec.~\ref{sec: input representation} and tri-modal alignment for dual-token generation in Sec.~\ref{sec: dual tokens}; ii) a shift-aware refinement module in Sec.~\ref{sec: shift anchor}; and iii) a memory-augmented mask decoder in Sec.~\ref{sec: decoder}.
Note that the decoder is flexible and can be replaced with diverse models (more details can be found in \textit{Appendix}~\ref{sec: ab on decoder}).
% Additional alternatives can be found in the supplementary materials.

\subsection{Input representation}
\label{sec: input representation}
% To facilitate tri-modal reasoning within the LLM, we employ a unified encoding and alignment strategy to map visual and auditory signals into the same semantic space with textual embeddings.
% Following LISA, the LLM helps generate the tokens for the segmentation prediction.

\noindent
\textbf{Visual encoding.}
Given a $T$-second video sequence, we utilize a pre-trained CLIP-ViT-L/14 \cite{radford2021clip} encoder to extract patch-level features.
For each frame, the encoder produces a feature map $f_v^t \in \mathbb{R}^{L_v \times D_v}$, where $L_v$ and $D_v$ represent the number of tokens and feature dimension, respectively.
The entire video features can be denoted as  $F_v = \{f_v^t\}_{t=1}^T$.
To compress these high-dimensional feature maps, we employ a Q-Former \cite{li2023blip2} with $V_q$ learnable queries to represent them as a compact set of visual tokens.
These tokens are subsequently projected into the LLM's input space via a two-layer multilayer perceptron (MLP).

\noindent
\textbf{Audio encoding.}
The corresponding audio stream is partitioned into $T$ segments $\{a_t\}_{t=1}^T$ synchronized with the video frames.
We leverage BEATs \cite{chen2023beats}, a self-supervised audio transformer, to capture rich semantic information from each segment, resulting in audio embeddings $f_a^t \in \mathbb{R}^{L_a \times D_a}$, where $L_a$ and $D_a$ represent the number of tokens and feature dimension, respectively.
Similar to the visual encoding, a dedicated Q-Former compresses these features into $A_q$ auditory queries.
Following an MLP projection layer, these queries are aligned with the textual modality and prepended to the LLM's input sequence alongside the visual and linguistic embeddings.

\subsection{MLLM for segmentation and dual-token generation}
\label{sec: dual tokens}
\noindent
\textbf{MLLM for Ref-AVS.}
Following the success of image-based reasoning segmentation models (\textit{e.g.}, LISA \cite{lai2024lisa}), recent Ref-AVS methods \cite{du2025crab, wang2025sam2love, ying2025oisa-1b} leverage the implicit reasoning of LLMs to generate a specialized segmentation token, denoted as \texttt{[seg]}.
In this paradigm, the MLLM processes the multimodal encodings to produce a high-level representation that drives a mask decoder $\mathrm{Dec}$.
A straightforward extension of this image-based approach can be formulated as:
\begin{equation}
   \texttt{[seg]}_{t=1}^N = \mathrm{MLLM}(V_q', A_q', Q_q'), \quad M_{t=1}^T = \mathrm{Dec}(F_v, \texttt{[seg]}_{t=1}^N),
\end{equation}
where $V_q'$, $A_q'$, and $Q_q'$ represent the projected embeddings from the video, audio, and text, respectively, and $N$ denotes the number of frames that an MLLM processes.
The MLLM outputs a sequence of frame-specific tokens to represent the target at each time step.
However, as previously analyzed, relying solely on independent \texttt{[seg]} tokens often fails to maintain a stable reference, making the model susceptible to target shift when sensory cues change.

\noindent
\textbf{Dual-token strategy.}
In contrast, we address the limitation by introducing a dual-token strategy designed to decouple dynamic per-frame focus from global semantic persistence.
Our method prompts the LLM to generate two distinct types of tokens: (1) temporal \texttt{[seg]} tokens: a sequence of tokens $\texttt{[seg]}_t \in \{\texttt{[seg]}_1, \dots, \texttt{[seg]}_N\}$, where each token is responsible for capturing the frame-specific tri-modal features for localized segmentation at time $t$;
(2) global \texttt{[query]} token: a singular token that serves as a persistent semantic anchor.
Unlike the \texttt{[seg]} tokens, the \texttt{[query]} token is optimized to summarize the long-term identity of the referred object across the entire video clip.
By generating these tokens in a unified causal chain, the MLLM facilitates structured reasoning over tri-modal alignment across vision, audio, and language.
The \texttt{[seg]} tokens provide the flexibility to capture temporal dynamics, while the \texttt{[query]} token provides a constant reference point.
This dual-token representation serves as the foundation for our subsequent shift-aware refinement, ensuring that even if a \texttt{[seg]} token drifts due to transient conflicts, the \texttt{[query]} anchor preserves the target's identity.

\subsection{Shift-aware refinement module}
\label{sec: shift anchor}
To maintain robust, persistent segmentation under dynamic conditions, we propose the shift-aware refinement module, which bridges the gap between the LLM's latent tokens and the frame-level modality evidence, ensuring the mask decoder receives a dynamically calibrated prompt.
Specifically, since raw visual and audio features reside in different modalities and often contain scene noise, we utilize the global semantic \texttt{[query]} to probe for target-specific evidence separately.
By treating the \texttt{[query]} token as a persistent anchor, we compute cross-modal attention \cite{vaswani2017attention} to filter out irrelevant background information and isolate the referent:
\begin{equation}
z_m^t = \mathrm{Softmax} \left( \frac{(W_q \texttt{[query]})(W_k f_m^t)^T}{\sqrt{d}} \right) V_{f_m^t}, \quad m \in {v, a},
\end{equation}
where $z_m^t$ represents the target-specific evidence within each modality at frame $t$.
Target shift is represented as the discrepancy between the LLM's current latent representation $\texttt{[seg]}_t$ and the actual observed evidence $z_m^t$.
We compute the modality-specific shifts as:
\begin{equation}
\Delta_m = \mathrm{MLP}(\phi(\texttt{[seg]}_t) - \psi(z_m^t)), \quad m \in {v, a},
\end{equation}
where $\phi$ and $\psi$ are projection layers.
These operations capture fine-grained visual motion or acoustic fluctuations that are inadequately represented by the LLM's discrete tokens, which tend to overlook subtle cues or overemphasize dominant modality signals.
% the LLM's discrete tokens overlook or over.
This estimation represents the ``shift'' because the deviation between the predicted semantic state and the synchronized sensory evidence signals a potential drift in focus, which the residual $\Delta_m$ explicitly quantifies.
To prevent the model from being misled by dominant-modality changes or transient noise, we introduce dynamic confidence gating for the estimated drifts.
Specifically, we compute the gating weights $\lambda_v$ and $\lambda_a$ based on the unnormalized attention logits generated during the evidence extraction stage.
We derive the global evidence intensity, $e_v$ and $e_a$, by averaging the pre-softmax alignment scores across the sequence length for each modality.
The modality-specific gating weights are then computed via a joint linear projection layer followed by a Sigmoid activation $\sigma$.
% Specifically, we compute the global average of the attention scores for each modality during the evidence stage, denoted as $e_v$ and $e_a$, and derive the modality-specific gating weights $\lambda_v$ and $\lambda_a$ via a joint linear projection layer followed by a Sigmoid activation $\sigma$.
% They are concatenated and passed through a joint linear projection layer to generate distinct gating vectors, which are followed by a Sigmoid activation $\sigma$.
We apply these gated weights to ensure that the refinement prioritizes the modality providing the most reliable evidence at any given time step, effectively filtering out unreliable signals from modality-specific changes.
It is formulated as:
\begin{equation}
\lambda_v, \lambda_a = \sigma(\mathrm{Linear}([e_v; e_a])), \quad \hat{\texttt{[seg]}_t} = \texttt{[seg]}_t + \mathrm{LN}(\lambda_v \Delta_v) + \mathrm{LN}(\lambda_a \Delta_a),
\end{equation}
where $\mathrm{LN}$ denotes LayerNorm.
The resulting $\hat{\texttt{[seg]}_t}$ token incorporates both the stable semantic anchor and the dynamic sensory evidence, providing the decoder with a shift-aware representation that is resilient to modality-specific drift.

\subsection{Mask decoding}
\label{sec: decoder}
% The final stage of our framework translates the shift-aware tokens into precise pixel-level masks while maintaining long-term identity consistency.
Unlike previous methods that either treat frames independently through propagation \cite{radman2025tsam} or employ standard SAM2 \cite{ravi2024sam2} as a generic temporal segmentation engine \cite{wang2025sam2love}, our mask decoder incorporates the object anchor and estimated shifts to maintain consistent and persistent perception of the target over time.
% process frames in isolation via propagation or employ standard SAM2 \cite{ravi2024sam2} as a powerful temporal engine for video segmentation, we leverage the previously obtained anchor and shifts to enhance the persistent perception of the target.
This step is a joint optimization of the memory bank from SAM2 and refined embeddings in an end-to-end manner.
Specifically, to bridge the gap between the LLM and the mask decoder, we locate the tokens and project the corresponding last layer's hidden states into the embedding space of the SAM2 prompt encoder.
Furthermore, we anchor the target with the \texttt{[query]} token alongside the temporal-level representations to enable inter-frame spatio-temporal consistency:
\begin{equation}
\mathcal{H}_t = \mathrm{MLP}(\hat{\texttt{[seg]}_t} + \alpha \texttt{[query]}),
\end{equation}
where $\mathcal{H}_t$ represents the prompt at frame $t$ and $\alpha$ denotes the fusion coefficient.
By utilizing the shift-refined tokens, this prompt inherently carries the corrected semantic anchor, effectively guiding the decoder to focus on the intended referent.
To ensure that the identity is preserved across the entire video sequence, we incorporate a continuous memory bank update.
For each frame $t$, the decoder generates the mask $M_t$ and simultaneously updates the temporal memory $\mathcal{M}$:
\begin{equation}
M_t= \mathrm{Dec}(\mathrm{Enc}_I(\mathcal{V}_t), \mathrm{Enc}_P(\mathcal{H}_t), \mathcal{M}),
\end{equation}
where $\mathrm{Enc}_I$ and $\mathrm{Enc}_P$ denote the SAM2 image encoder and prompt encoder, respectively.
The shift-aware prompting and memory-based propagation further allow SAIL to achieve robust spatio-temporal identity consistency throughout the video timeline.

\section{Experiments}
\label{sec: experiments}

\subsection{Datasets and evaluation metric}
\label{sec: dataset}
\noindent
\textbf{Dataset.}
The experiments were primarily conducted on the widely-used Ref-AVSBench \cite{wang2024refavs}.
% , a comprehensive dataset designed for Ref-AVS.
It provides a dense mapping of 20,261 linguistic queries to pixel-level annotations across 4,002 diverse video sequences.
% Ref-AVSBench is distinguished by its cross-modal complexity, requiring models to fuse audio--visual cues under textual guidance.
The benchmark follows a standard split of 2,908 training, 276 validation, and 818 testing videos.
Notably, the test set is further divided into three distinct scenarios:
i) \textit{Seen} (39 known categories);
ii) \textit{Unseen} (13 novel categories for testing out-of-distribution robustness);
iii) \textit{Null} (scenarios involving non-existent or invisible targets).
% The last serves as a critical diagnostic to evaluate the model's susceptibility to ``hallucinating'' segmented regions when a valid target is missing.
We also evaluated models on $\mathrm{R^2}$-AVSBench \cite{zhou2025tgs}, an extension of Ref-AVSBench that incorporates reasoning-focused enhancements and comprises 400 testing videos (more details can be found in \textit{Appendix}~\ref{sec: more detail}).
% More details can be found in the \textit{supplementary materials}.
% and provided discussion on the compared results in Appendix Sec.~\ref{sec: avqa}.

\noindent
\textbf{Evaluation metric.}
Following the standard protocol in previous Ref-AVS studies \cite{wang2024refavs, wang2025sam2love}, we report the Jaccard index ($\mathcal{J}$) and F-score ($\mathcal{F}$) scores to quantify segmentation quality.
For the \textit{Null} subset, we report the S-score ($\mathcal{S}$), defined as the ratio of predicted mask area to total background area.
This specialized metric assesses the model's ability to suppress false positives.
% a lower $\mathcal{S}$ value demonstrates a higher capacity for rejecting invalid objects.

\subsection{Implementation details}
\label{sec: details}
Videos were segmented into 1-second clips from 10-second recordings.
% Visual features were extracted using CLIP-ViT-L/14 \cite{radford2021clip}, and audio features were captured via BEATs \cite{chen2023beats}.
For both vision and audio modalities, we employed a Q-Former \cite{li2023blip2} with 32 learnable query tokens to compress the features before projection.
% features were projected into 512-dimensional vectors for consistency.
We adopted Chat-UniVi-7B \cite{jin2024chatunivi} as our MLLM and performed fine-tuning using LoRA \cite{hu2022lora} with a rank of 8.
We bypassed the need for auxiliary data for projector pre-training by jointly training the alignment layers and the memory-augmented SAM2 (Hiera-L) \cite{ravi2024sam2} decoder in a single, data-efficient end-to-end stage.
We trained the model using AdamW \cite{loshchilov2017adamw} optimizer with an initial learning rate of 1\textit{e}-4 and a cosine decay schedule.
All experiments were conducted using PyTorch on two NVIDIA H100 GPUs.
% The model was trained for 10 epochs with a batch size of 2 per device.
% The hyper-parameters $\lambda_{txt}$, $\lambda_{bce}$, $\lambda_{dice}$ and $\alpha$ were set to 1.0, 1.0, 0.5 and 0.1, respectively.
More details are provided in \textit{Appendix}~\ref{sec: more detail}.
% Training details such as gpus, epochs, batch size, and learning rate are provided in the \textit{supplementary materials}.

\subsection{Comparison with state-of-the-art methods}
\label{sec: comparison}
\noindent
\textbf{Compared methods.}
To validate the effectiveness of the proposed method, we compared it with previous state-of-the-art approaches across three related tasks: AVS, Ref-VOS, and Ref-AVS.
For AVS, we compared against AVSBench \cite{zhou2022avs}, AVSegFormer \cite{gao2024avsegformer}, GAVS \cite{wang2024gavs}, and SAMA \cite{liu2024sama}.
We further adapted Ref-VOS methods that rely solely on text and vision, such as ReferFormer \cite{wu2022referformer} and R2-VOS \cite{li2023r2vos}.
Finally, we compared against a comprehensive set of Ref-AVS methods, including specialized multimodal methods such as EEMC \cite{wang2024refavs}, TSAM \cite{radman2025tsam}, and SAM2-LOVE \cite{wang2025sam2love}, as well as MLLM-based models including Crab \cite{du2025crab}, OISA-1B \cite{ying2025oisa-1b}, TGS-Agent \cite{zhou2025tgs}, and AURORA \cite{luo2025aurora}.

\begin{table}[t]\scriptsize{}
  \caption{Comparison with existing methods on Ref-AVSBench \cite{wang2024refavs}, where Mix represents the mean performance of Seen and Unseen subsets.
  % reporting performance across Seen, Unseen, Mix, and Null test splits, where Mix represents the mean performance of Seen and Unseen subsets.
  % Results for existing methods are sourced from either SAM2-LOVE \cite{wang2025sam2love} or their original works.
  $\dagger$ denotes methods requiring additional data for instruction tuning; $\ddagger$ indicates models that further necessitate external components in multiple stages.
  The best results are \textbf{bold} while the second best are \underline{underlined}.
  }
  % \vspace{-2mm}

  % \textbf{$\mathcal{J}$}\&\textbf{$\mathcal{F}$}
  \centering
  \setlength\tabcolsep{3pt}
  \begin{tabular}{p{3.cm}|c|c|cc|cc|cc|c}
  \toprule
  \multirow{2}*{Method} & \multirow{2}*{Task} & \multirow{2}*{Model} & \multicolumn{2}{c|}{Seen$\uparrow$} & \multicolumn{2}{c|}{Unseen$\uparrow$} & \multicolumn{2}{c|}{Mix$\uparrow$} & \multicolumn{1}{c}{Null$\downarrow$}\\
  % \cmidrule(lr){4-5} \cmidrule(lr){6-7} \cmidrule(lr){8-9}
  
 & &  & \textbf{$\mathcal{J}$} & \textbf{$\mathcal{F}$} & \textbf{$\mathcal{J}$} & \textbf{$\mathcal{F}$}  & \textbf{$\mathcal{J}$} & \textbf{$\mathcal{F}$}  & \textbf{$\mathcal{S}$}\\
 \midrule
  AVSBench \cite{zhou2022avs} \tiny{\textit{eccv22}} & \multirow{4}{*}{AVS} & PVT-v2 & 23.2 & 51.1 & 32.4 & 54.7 & 27.8 & 52.9 & 0.208\\
  AVSegFormer \cite{gao2024avsegformer} \tiny{\textit{aaai24}} & & PVT-v2 & 33.5 & 47.0 & 36.1 & 50.1 & 34.8 & 48.6  & 0.171\\
  GAVS \cite{wang2024gavs} \tiny{\textit{aaai24}} & & SAM & 28.9 & 49.8 & 29.8 & 49.7 & 29.4 & 49.8 & 0.190\\
  SAMA \cite{liu2024sama} \tiny{\textit{wacv24}} & & SAM & 39.2 & 56.2 & 47.5 & 56.6 & 43.4 & 56.4 &  0.130\\

  \hline
  ReferFormer \cite{wu2022referformer} \tiny{\textit{cvpr22}} & \multirow{2}{*}{Ref-VOS} &  Video-Swin & 31.3 & 50.1 & 30.4 & 48.8 & 30.9 & 49.5 & 0.176 \\
  R2-VOS \cite{li2023r2vos} \tiny{\textit{iccv23}} &  &  Video-Swin & 25.0 & 41.0 & 27.9 & 49.8 & 26.5 & 45.4 & 0.183\\
  
  \hline
  EEMC \cite{wang2024refavs}  \tiny{\textit{eccv24}} & \multirow{7}{*}{Ref-AVS} & Mask2Former & 34.2 & 51.3 & 49.5 & 64.8 & 41.9 & 58.1 & \textbf{0.007}\\
  TSAM \cite{radman2025tsam} \tiny{\textit{cvpr25}} & & SAM & 43.4 & 56.8 & 54.6 & 66.4 & 49.0 & 61.6 & 0.017\\
  SAM2-LOVE \cite{wang2025sam2love} \tiny{\textit{cvpr25}} & & SAM2 & 43.5 & 51.9 & 66.5 & 72.3 & 55.0 & 62.1 & 0.230\\
  Crab \cite{du2025crab} $\dagger$ \tiny{\textit{cvpr25}} & & SAM + LLaMA2 & 40.5 & - & 45.6 & - & 43.1 & - & -\\
  OISA-1B \cite{ying2025oisa-1b} $\dagger$ \tiny{\textit{iccv25}} & & Mask2Former + InternVL2 & 51.7 & 58.7 & 58.3 & 65.1 & 54.5 & 61.4 & 0.098\\
  TGS-Agent \cite{zhou2025tgs} $\ddagger$ \tiny{\textit{aaai26}} & & SAM2 + Groundin-DINO + LLaMA2 & 49.5 & 60.4 & \textbf{73.2} & \textbf{80.6} & 61.3 & 70.5 & 0.035\\
  AURORA \cite{luo2025aurora} $\ddagger$ \tiny{\textit{aaai26}} & & SAM + Qwen + VideoLLaMA2  & \underline{63.2} & \underline{72.8} & 69.7 & 76.4 & \underline{66.5} & \underline{74.6} & -\\
  \hline
  SAIL (ours) & Ref-AVS & SAM2 + Chat-UniVi & \textbf{68.4} & \textbf{77.5} & \underline{72.0} & \underline{79.5} & \textbf{70.2} & \textbf{78.5} & \underline{0.010}\\
  \bottomrule
  \end{tabular}
  \label{tab: comparison with SOTA for ref-avs}
  \vspace{-5mm}
\end{table}

\noindent
\textbf{Quantitative results.}
The quantitative results on the Ref-AVSBench dataset are reported in Table~\ref{tab: comparison with SOTA for ref-avs}.
The specific components of these methods are listed in this table (further details are provided in \textit{Appendix}~\ref{sec: more detail}).
Our proposed SAIL achieved state-of-the-art performance across most metrics, significantly outperforming existing AVS, Ref-VOS, and Ref-AVS methods.
% Compared to methods designed for general AVS (\textit{e.g.}, AVSBench, SAMA) or language-guided video segmentation (\textit{e.g.}, ReferFormer, R2-VOS), SAIL demonstrated substantial gains across all metrics, which are consistent with trends observed in other Ref-AVS models.
For instance, on the Mix split, SAIL surpassed the best-performing AVS baseline, SAMA, and representative Ref-VOS baseline, ReferFormer, by 26.8\% and 39.3\% in $\mathcal{J}$, respectively.
These significant improvements underscore that simply extending single-modality or generic segmentation frameworks is insufficient for the complex tri-modal reasoning required in audio--visual scenarios.
Notably, SAIL established a new performance ceiling for Ref-AVS.
% particularly in the Seen and Mix splits.
For instance, SAIL achieved 78.5\% $\mathcal{F}$ on Mix, compared to 58.1\% by the baseline EEMC and 61.4\% by the omni-model OISA-1B.
It also outperformed the previous SOTA method, AURORA, by 3.7\% $\mathcal{J}$ and 3.9\% $\mathcal{F}$.
Detailed analysis shows that on the Seen split, our method achieved a 5.2\% gain over AURORA in $\mathcal{J}$.
% , while performance in the Unseen split was improved by 2.3\%.
% These results demonstrate the strength of our shift-aware modeling design.
These gains arise because SAIL explicitly addresses the target shift problem through token refinement rather than relying solely on the implicit reasoning of LLMs.
Despite not leveraging additional instruction-tuning data (as used in OISA-1B) or complex multi-stage pipelines (as used in TGS-Agent and AURORA) with external components such as Grounding-DINO \cite{liu2024grounding} and Qwen \cite{xu2025qwen3}), SAIL still performed competitively on the Unseen split.
It trailed the best-performing model, TGS-Agent, by only 1.1\% in $\mathcal{F}$, while significantly outperforming other specialized multimodal methods (\textit{e.g.,} SAM2-LOVE +7.2\%) and MLLM-based methods (\textit{e.g.,} AURORA +3.1\%).
The effectiveness of SAIL was further highlighted by the results on the Null split.
% which evaluates the model's ability to refrain from segmenting when the target is absent.
SAIL achieved a remarkably low $\mathcal{S}$ score of 0.010, trailing only EEMC (0.007) but significantly outperforming other MLLM-based competitors like OISA-1B (0.098).
This suggests that our method with the dual-token strategy effectively suppresses false positives caused by sensory noise.
To further validate the effectiveness of our method in addressing the target shift problem, we curated a specialized evaluation subset from the Ref-AVSBench test split, specifically targeting samples prone to significant temporal changes.
Due to space constraints, a detailed description of this subset and an extended comparative discussion are provided in \textit{Appendix}~\ref{sec: additional evaluation}.

\noindent
\textbf{Evaluation on $\mathrm{R^2}$-AVSBench.}
To further assess the reasoning capabilities and generalizability of our model, we evaluated it on the $\mathrm{R^2}$-AVSBench dataset.
This benchmark replaces explicit object descriptions with implicit reasoning-heavy instructions, significantly increasing the difficulty of the task.
The quantitative results are reported in Table~\ref{tab: comparison with SOTA for r2}.
Compared to the results on Ref-AVSBench, we observe a general performance decline across all models, which is consistent with the increased complexity of interpreting implicit linguistic cues.
Notably, SAIL maintained the inherent reasoning advantages of the MLLM-based architecture and achieved robust segmentation performance through our specialized anchor and shift-aware designs.
Overall, SAIL achieved 64.4\% in $\mathcal{J}\&\mathcal{F}$ on the Mix split, outperforming the previous SOTA method, TGS-Agent, by 4.3\%.
% Although TGS-Agent achieved higher scores on the Unseen split due to its reliance on the external Grounding-DINO, SAIL demonstrated superior balance and robustness across the benchmark.
While TGS-Agent achieved higher performance on the Unseen split, it relied on a multi-stage pipeline with external components (\textit{e.g.}, LLM-based reasoning and Grounding-DINO for explicit object localization), which may offer advantages in object grounding.
In contrast, SAIL adopted a unified end-to-end framework without such auxiliary modules, while maintaining competitive performance and stronger robustness across the benchmark.
These results indicate that even when linguistic inputs are implicit, our framework effectively grounds the referred object by leveraging the refined tri-modal evidence, demonstrating strong reasoning ability and generalization.

\begin{figure}[t]
    \centering
    \includegraphics[width=0.85\linewidth]{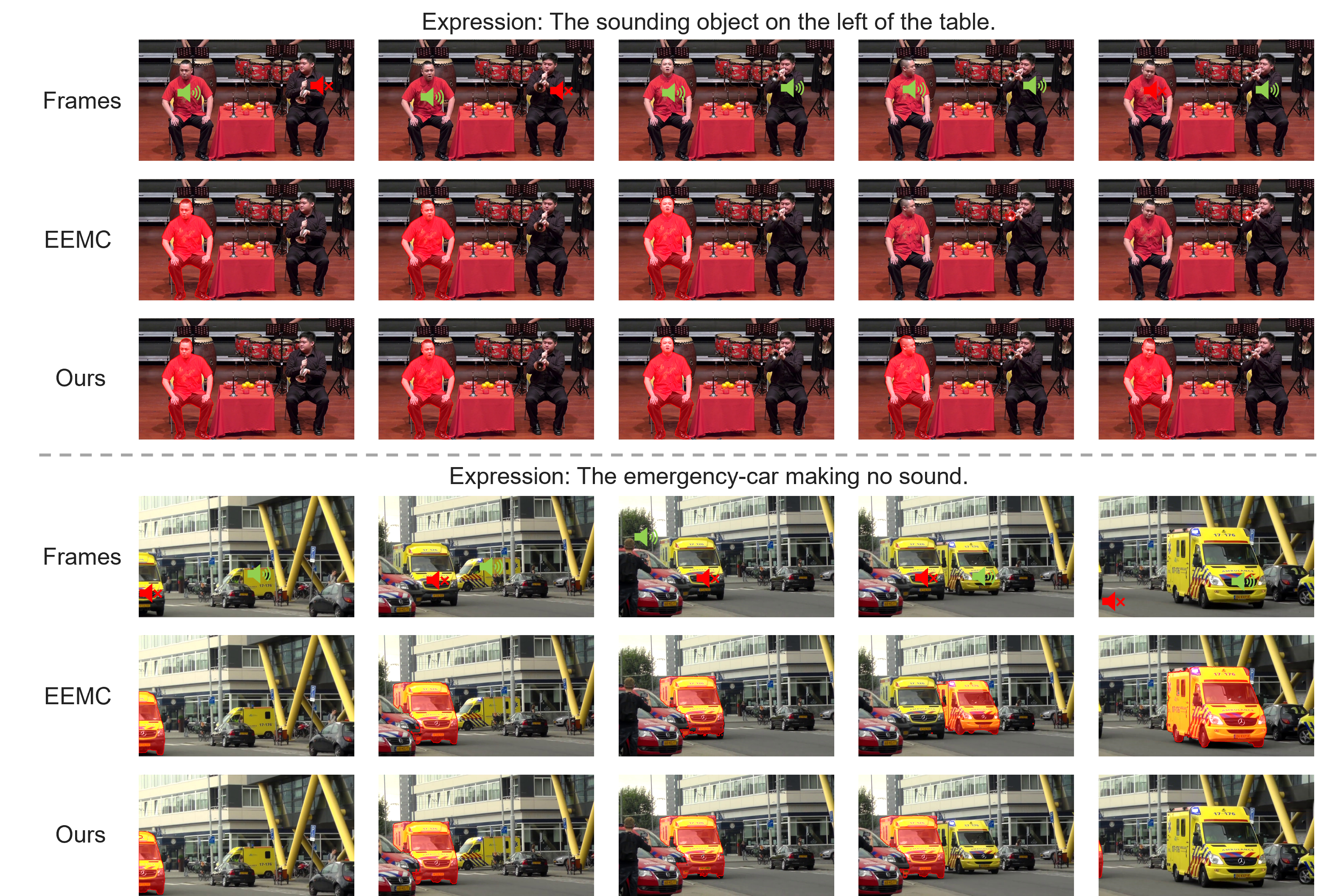}
    \caption{Qualitative results of SAIL and baseline from Ref-AVSBench \cite{wang2024refavs}.
    % The top row illustrates a case from the Seen split, while the bottom row showcases one from the Unseen split.
    % \textcolor{red}{TODO}.
    }
    \label{fig: qualitative}
    \vspace{-5mm}
\end{figure}

\noindent
\textbf{Qualitative results.}
In Fig.~\ref{fig: qualitative}, we present qualitative results from the Ref-AVS test set.
Overall, SAIL demonstrates superior target identification and segmentation precision.
In the top example, while EEMC initially segmented the man performing vocal mimicry, it lost focus as the sound fades and other instruments (\textit{e.g.}, suona) begin to play.
In contrast, SAIL consistently identified the referent by leveraging shift-aware refinement to mitigate the impact of transient auditory changes.
In the bottom example involving transient occlusion, EEMC erroneously tracked a distractor of similar appearance.
SAIL successfully captured the latent shift even when the distractor hides behind and reappears near the true referent, effectively distinguishing between two similar emergency-cars.
Notably, even when only a small portion of the target is visible in the final frames, our method maintained a correct and robust segmentation.
These examples highlight SAIL's ability to leverage tri-modal dynamics and global anchoring, enabling accurate reasoning and segmentation in complex audio--visual scenes.
Additional qualitative results and extended discussions are provided in \textit{Appendix}~\ref{sec: more vis}.
% Due to space constraints, additional qualitative results, comprehensive comparisons with baseline methods, and extended discussions are provided in \textit{Appendix}~\ref{sec: more vis}.

\begin{table}[t]\scriptsize{}
  \caption{Comparison on $\mathrm{R^2}$-AVSBench \cite{zhou2025tgs}.
  Results for existing methods are sourced from either TGS-Agent or our re-implementations (in $\ast$).
  Note that TGS-Agent employed external Grounding-DINO \cite{liu2024grounding} for explicit object localization.
  $\mathcal{J}$\&$\mathcal{F}$ represents the average of $\mathcal{J}$ and $\mathcal{F}$.
  }
  \centering
  \setlength\tabcolsep{4pt}
  \begin{tabular}{p{3.5cm}|ccc|ccc|ccc}
  \toprule
  \multirow{2}*{Method} & \multicolumn{3}{c|}{Seen$\uparrow$} & \multicolumn{3}{c|}{Unseen$\uparrow$} & \multicolumn{3}{c}{Mix$\uparrow$}\\
  % \cmidrule(lr){4-5} \cmidrule(lr){6-7} \cmidrule(lr){8-9}
  
  & \textbf{$\mathcal{J}$} & \textbf{$\mathcal{F}$} & \textbf{$\mathcal{J}$}\&\textbf{$\mathcal{F}$} & \textbf{$\mathcal{J}$} & \textbf{$\mathcal{F}$}  & \textbf{$\mathcal{J}$}\&\textbf{$\mathcal{F}$} & \textbf{$\mathcal{J}$} & \textbf{$\mathcal{F}$}  & \textbf{$\mathcal{J}$}\&\textbf{$\mathcal{F}$}\\
 \midrule
  EEMC \cite{wang2024refavs}  \tiny{\textit{eccv24}} & 22.6 & 39.9 & 31.2 & 38.7 & 56.9 & 47.8 &  30.6 & 48.4 & 39.5\\
  Grounded-SAM2 \cite{ren2024groundedsam2}  \tiny{\textit{arxiv24}} & 21.4 & 30.1 & 25.7 & 44.9 & 53.4 & 49.1 & 33.2 & 41.7 & 37.4\\
  Crab \cite{du2025crab} \tiny{\textit{cvpr25}} & 19.6 & 32.1 & 25.9 & 36.9 & 52.0 & 44.4 & 28.2 & 42.1 & 35.2\\
  TGS-Agent \cite{zhou2025tgs} \tiny{\textit{aaai26}} & 42.7 & 52.3 & 47.5 & \textbf{68.3} & \textbf{77.0} & \textbf{72.7} & \underline{55.5} & \underline{64.7} & \underline{60.1}\\
  AURORA \cite{luo2025aurora} $\ast$ \tiny{\textit{aaai26}} & \underline{50.5} & \underline{61.1} & \underline{55.8} & 55.0 & 64.4 & 59.7 & 52.8  & 62.8 & 57.8\\
  \hline
  SAIL (ours) & \textbf{58.3} & \textbf{68.9} & \textbf{63.6} & \underline{60.2} & \underline{70.1} & \underline{65.2} & \textbf{59.3} & \textbf{69.5} & \textbf{64.4}\\
  \bottomrule
  \end{tabular}
  \label{tab: comparison with SOTA for r2}
  \vspace{-5mm}
\end{table}

\begin{table}[tb]\scriptsize{}
  \begin{minipage}[t]{0.55\linewidth}
    \centering
\caption{Ablation study on the main designs.
The dual-token strategy, shift-aware refinement, and anchor alignment are denoted as DK, SR, and AA, respectively.
% (in the decoder)
}
\scalebox{0.90}{
\setlength\tabcolsep{4pt}
\begin{tabular}{l|cc|cc|cc|c}
\toprule
\multirow{2}{*}{Method} & \multicolumn{2}{|c}{Seen$\uparrow$} & \multicolumn{2}{|c}{Unseen$\uparrow$} & \multicolumn{2}{|c|}{Mix$\uparrow$} & \multicolumn{1}{c}{Null$\downarrow$}\\
 & \textbf{$\mathcal{J}$} & \textbf{$\mathcal{F}$} & \textbf{$\mathcal{J}$} & \textbf{$\mathcal{F}$} & \textbf{$\mathcal{J}$} & \textbf{$\mathcal{F}$} & \textbf{$\mathcal{S}$}\\
 \midrule
 Baseline & 41.3 & 55.6 & 45.4 & 61.3 & 43.4 & 58.5 & 0.049\\
 +DK & 50.7 & 59.2 & 55.1 & 67.0 & 52.9 & 63.1 & 0.037\\
 +DK+SR & 61.8 & 69.7 & 68.3 & 75.5 & 65.1 & 72.6 & 0.019\\
 +DK+SR+AA & 64.6 & 73.8 & 70.2 & 77.1 & 67.4 & 75.5 & 0.015\\
 +DK+SR+$\mathcal{M}$ & 65.0 & 75.2 & 70.4 & 78.0 & 67.7 & 76.6 & 0.013\\
 \hline
 SAIL (ours) & \textbf{68.4} & \textbf{77.5} & \textbf{72.0} & \textbf{79.5} & \textbf{70.2} & \textbf{78.5} & \textbf{0.010}\\
\bottomrule
\end{tabular}
}
  \label{tab: ablation on main moduels}
  \end{minipage}
  \hfill
  \begin{minipage}[t]{0.42\linewidth}
    \centering
\centering
\caption{Ablation study on the alternative design strategies (Pre-LLM, In-LLM, and Post-LLM) for tackling the target shift problem.
% For space, the explicit object embeddings, implicit shift representation, and learnable referent query are denoted as Object-emb, Shift-rep and Ref-query, respectively.
% (pre-LLM) (in-LLM)  (post-LLM)
}
\scalebox{0.98}{
\setlength\tabcolsep{2pt}
\begin{tabular}{l|cc|cc|cc|c}
\toprule
\multirow{2}{*}{Method} & \multicolumn{2}{|c}{Seen$\uparrow$} & \multicolumn{2}{|c}{Unseen$\uparrow$} & \multicolumn{2}{|c|}{Mix$\uparrow$} & \multicolumn{1}{c}{Null$\downarrow$}\\
 & \textbf{$\mathcal{J}$} & \textbf{$\mathcal{F}$} & \textbf{$\mathcal{J}$} & \textbf{$\mathcal{F}$} & \textbf{$\mathcal{J}$} & \textbf{$\mathcal{F}$} & \textbf{$\mathcal{S}$}\\
 \midrule
 Pre-LLM & 43.8 & 51.7 & 58.0 & 69.5 & 50.9 & 60.6 & 0.075\\
 In-LLM & 66.3 & 74.1 & 59.7 & 66.2 & 63.0 & 70.2 & 0.042 \\
 Post-LLM & 55.7 & 61.5 & 63.8 & 70.6 & 59.8 & 66.1 & 0.061 \\
 \hline
 SAIL (ours) & \textbf{68.4} & \textbf{77.5} & \textbf{72.0} & \textbf{79.5} & \textbf{70.2} & \textbf{78.5} & \textbf{0.010}\\
\bottomrule
\end{tabular}
}
\label{tab: ablation on shift-aware}
  \end{minipage}
  \vspace{-5mm}
\end{table}

\subsection{Ablation studies}
\label{sec: ablation}

\noindent
\textbf{Ablation study on main components.}
To verify the effectiveness of the main components in SAIL, we conducted a cumulative ablation study on the Ref-AVSBench dataset.
% by progressively integrating our core designs.
Table~\ref{tab: ablation on main moduels} reports the corresponding results, starting from a standard LISA-style MLLM baseline and progressively integrating our core designs.
Removing all proposed designs resulted in a substantial performance drop of 20\% in $\mathcal{F}$ on the Mix split, highlighting the overall importance of our framework.
Replacing the single segmentation token with our dual-token strategy (\textit{i.e.,} \texttt{[seg]} and \texttt{[query]}) yielded a significant boost, suggesting that decoupling frame-specific dynamics from a persistent semantic anchor helps the model maintain a consistent target identity.
The introduction of the shift-aware refinement module provided the most substantial gain, increasing $\mathcal{J}$ and $\mathcal{F}$ by 12.2\% and 9.5\% on Mix, respectively.
These sharp improvements validate our hypothesis that latent space drift, caused by modality noise or scene changes, is a primary bottleneck in Ref-AVS.
By explicitly estimating and rectifying the shift between the LLM's tokens and target-specific evidence, the model achieved much higher grounding precision.
Furthermore, aligning the weighted anchor with the segmentation representation and incorporating the SAM2 memory bank update (\textit{i.e.,} $\mathcal{M}$) both independently enhanced temporal stability.
For example, the anchor alignment improved $\mathcal{J}$ by 2.3\% on the Mix split, while the memory bank update provided a 2.6\% boost.
These components ensure the referent is not ``forgotten'' during mask prediction.
Finally, the complete SAIL framework, which integrates all modules, achieved peak performance across all splits and a minimum $\mathcal{S}$ score of 0.010 on the Null split, effectively suppressing false positives while maximizing segmentation accuracy.
More ablation studies on these modules and their alternatives can be found in \textit{Appendix}~\ref{sec: additional abl}.

\noindent
\textbf{Ablation on alternative design strategies for handling target shift.}
To further validate our shift-aware refinement designs, we compared it against three alternative strategies for tackling the target shift problem.
Table~\ref{tab: ablation on shift-aware} reports the corresponding results.
(1) \textit{Pre-LLM}:
Following standard practices in video object tracking and segmentation \cite{fu2025objectrelator, wang2025objectvqa}, we extracted explicit object-level features via a pre-trained detector and compressed them as auxiliary LLM inputs.
While providing concrete visual priors, this strategy yielded the lowest performance (\textit{e.g.}, 50.9\% in $\mathcal{J}$), demonstrating that pre-LLM injection is insufficient for dynamic tri-modal reasoning.
(2) \textit{In-LLM}:
Inspired by prompt-learning techniques \cite{lai2024lisa, yan2024visa}, we generated an implicit shift token alongside the segmentation token within the LLM.
This strategy improved performance on the Seen split (74.1\% in $\mathcal{F}$) but suffered on the Unseen split (66.2\% in $\mathcal{F}$), suggesting that the LLM struggles to implicitly model spatial-temporal drift without explicit geometric grounding.
(3) \textit{Post-LLM}:
Drawing from instructional editing models \cite{yu2025veggiepost}, we used a learnable query token to implicitly extract object-level dynamics from video features, which was then fused with the $\texttt{[seg]}$ tokens.
This strategy also failed to match SAIL, as it lacks the global semantic anchoring required for consistent Ref-AVS.
In contrast, our SAIL framework achieved the best results across all splits.
This confirms that rectifying latent shift with target-specific evidence is more effective than relying on explicit pre-processing or purely implicit token representations (more details are provided in \textit{Appendix}~\ref{sec: ab on alternative}).
% More  the \textit{supplementary materials}.

% \subsection{Footnotes}

% Footnotes should be used sparingly.  If you do require a footnote, indicate
% footnotes with a number\footnote{Sample of the first footnote.} in the
% text. Place the footnotes at the bottom of the page on which they appear.
% Precede the footnote with a horizontal rule of 2~inches (12~picas).

% Note that footnotes are properly typeset \emph{after} punctuation
% marks.\footnote{As in this example.}

% \subsection{Qualitative Results}
% \label{sec: qualitative}

\section{Conclusion}
\label{sec: conclusion}
In this work, we explore the challenging problem of target shift in referring audio--visual segmentation and propose SAIL, a novel MLLM-based framework.
To achieve robust and persistent segmentation in tri-modal video reasoning, we introduce a dual-token design that decouples frame-specific dynamics from a temporally consistent object identity. Building on this, we develop a shift-aware refinement module that explicitly estimates and rectifies modality-induced latent drift.
By integrating the refined representations into a SAM2-based mask decoder, SAIL enables precise and temporally consistent segmentation.
Extensive experiments on Ref-AVSBench and $\mathrm{R^2}$-AVSBench demonstrate that SAIL consistently outperforms state-of-the-art methods, while ablation studies confirm the effectiveness of its components.
We believe this work highlights the importance of identity-consistent representation learning and offers a promising direction for robust audio--video understanding.

{
\small
\bibliographystyle{plainnat}
\bibliography{neurips}
}
% \section*{References}

% References follow the acknowledgments in the camera-ready paper. Use unnumbered first-level heading for
% the references. Any choice of citation style is acceptable as long as you are
% consistent. It is permissible to reduce the font size to \verb+small+ (9 point)
% when listing the references.
% Note that the Reference section does not count towards the page limit.
% \medskip

% {
% \small

% [1] Alexander, J.A.\ \& Mozer, M.C.\ (1995) Template-based algorithms for
% connectionist rule extraction. In G.\ Tesauro, D.S.\ Touretzky and T.K.\ Leen
% (eds.), {\it Advances in Neural Information Processing Systems 7},
% pp.\ 609--616. Cambridge, MA: MIT Press.

% [2] Bower, J.M.\ \& Beeman, D.\ (1995) {\it The Book of GENESIS: Exploring
%   Realistic Neural Models with the GEneral NEural SImulation System.}  New York:
% TELOS/Springer--Verlag.

% [3] Hasselmo, M.E., Schnell, E.\ \& Barkai, E.\ (1995) Dynamics of learning and
% recall at excitatory recurrent synapses and cholinergic modulation in rat
% hippocampal region CA3. {\it Journal of Neuroscience} {\bf 15}(7):5249-5262.
% }

%%%%%%%%%%%%%%%%%%%%%%%%%%%%%%%%%%%%%%%%%%%%%%%%%%%%%%%%%%%%

\appendix
\section{Appendix}
This supplementary document provides comprehensive details regarding the datasets and model architectures (Sec.~\ref{sec: more detail}).
We further present an expanded evaluation of target shifts (Sec.~\ref{sec: additional evaluation}), additional ablation studies (Sec.~\ref{sec: additional abl}), and more qualitative visualizations (Sec.~\ref{sec: more vis}).
We also provide complexity analysis (Sec.~\ref{sec: computation}) and a discussion (Sec.~\ref{sec: discussion}) for potential limitations and future work.

\subsection{More details on datasets and models}
\label{sec: more detail}

\subsubsection{Reasoning capability and dataset}
While Ref-AVSBench \cite{wang2024refavs} establishes the foundational benchmark for Ref-AVS, its linguistic queries primarily rely on explicit descriptions (\textit{e.g.}, ``The person playing the instrument'').
In contrast, $\mathrm{R^2}$-AVSBench \cite{zhou2025tgs} is specifically designed to evaluate more sophisticated cognitive abilities by introducing implicit reasoning challenges.
The core distinction of $\mathrm{R^2}$-AVSBench lies in its query formulation.
Leveraging Gemini \cite{team2024gemini}, the dataset replaces direct nouns with functional or acoustic descriptions, significantly increasing lexical and structural diversity.
For example, a simple query in Ref-AVSBench such as ``The dog barking in the yard'' is transformed in $\mathrm{R^2}$-AVSBench into ``The creature making a repetitive high-pitched alert sound while standing on the grass.''
This change requires the model to not only localize the object but to first reason about the relationship between the acoustic signatures and the visual entities.

Consequently, $\mathrm{R^2}$-AVSBench provides an extended evaluation set (400 in total: 220 and 180 videos for Seen and Unseen splits) derived from the Ref-AVSBench test split.
The increased complexity is reflected in the linguistic statistics: the average query length in $\mathrm{R^2}$-AVSBench is 11.73 words, a significant expansion compared to the 7.08 words found in the original benchmark.
% Overall, the $\mathrm{R^2}$-AVSBench is particularly relevant for validating our SAIL framework.
% Our dual-token strategy is uniquely suited for this task: the \texttt{[query]} token acts as a stable semantic anchor that decodes complex implicit intent, while the \texttt{[seg]} token handles the dynamic localization, effectively preventing the model from losing the target.
The experimental results reported in the main text (Sec.~\ref{sec: comparison}) further validate the superior reasoning capabilities of our proposed framework in handling complex, implicit Ref-AVS scenarios.

\subsubsection{Architectural details of compared methods}
% In the main text, we briefly introduce the compared methods in Sec. 4.3 and alternative methods in Sec. 4.4.
% Here we provide more details on the architectures of these methods.

In this section, we provide extended technical details for the baseline and state-of-the-art methods discussed in the main text (Sec.~\ref{sec: comparison}), focusing on their underlying architectural paradigms:
\begin{itemize}[leftmargin=*, itemsep=0pt]
\item AVSBench \cite{zhou2022avs}.
It employs a transformer-based architecture with a PVT-v2 \cite{wang2022pvtv2} backbone and introduces a temporal pixel-level grouping strategy to associate audio features with spatial visual maps for sounding object localization.

\item AVSegFormer \cite{gao2024avsegformer}.
Built on the PVT-v2 backbone, this model utilizes a transformer decoder where audio cues are injected into visual queries via cross-attention to amplify global audio--visual context and refine mask boundaries.

\item GAVS \cite{wang2024gavs}. It integrates the Segment Anything Model (SAM \cite{kirillov2023sam}) into the AVS task using a semantic-aware audio prompting strategy, guiding the foundation model to focus specifically on sounding regions.

\item SAMA \cite{liu2024sama}. It is an evolution of SAM-based AVS that introduces multimodal attention. It prompts SAM with localized audio--visual fused features to leverage its high-fidelity mask generation capabilities.

\item ReferFormer \cite{wu2022referformer}. It is a unified framework for Ref-VOS using a Video-Swin \cite{liu2022videoswin} backbone. It treats linguistic expressions as object queries that directly attend to the most relevant spatial regions across video frames.

\item R2-VOS \cite{li2023r2vos}. Utilizing Video-Swin, this method introduces a structural text-to-text cycle consistency constraint. It jointly models segmentation and text reconstruction to enforce semantic consensus between video--text pairs.

\item EEMC \cite{wang2024refavs}. Built on Mask2Former \cite{cheng2022mask2former}, this model proposes an expression enhancement module that mines multimodal cues. It focuses on integrating dynamic audio--visual information directly into the reference expressions to resolve alignment.

\item TSAM \cite{radman2025tsam}. It is an end-to-end SAM-based architecture that redefines visual prompts as data-driven multimodal prompts. It incorporates a temporal modeling branch with sparse and dense prompting modules to stabilize segmentation.

\item SAM2-LOVE \cite{wang2025sam2love}. It is the first adaptation of SAM2 \cite{ravi2024sam2} for Ref-AVS. It utilizes a multimodal fusion module to compress all the modalities' information into a learnable token, aligning with the SAM2 for segmenting the referred target. 

\item Crab \cite{du2025crab}. It combines LLaMA-2-7b-chat \cite{touvron2023llama} with SAM and introduces an interaction-aware LoRA \cite{hu2022lora} structure. It leverages LLM reasoning to interpret complex instructions but requires extensive auxiliary instruction-tuning data.

\item OISA-1B \cite{ying2025oisa-1b}. It employs InternVL2-1B \cite{chen2024internvl2} and Mask2Former, emphasizing large-scale vision--language pre-training to improve open-vocabulary recognition and handle omnimodal reasoning tasks. 

\item TGS-Agent \cite{zhou2025tgs}. It is a modular ``Think-Ground-Segment'' pipeline coordinating LLaMA-2-7b-chat, Grounding-DINO \cite{liu2024grounding}, and SAM2. It uses the LLM as an agent to manage independent expert models for explicit description generation, grounding, and segmentation.

\item AURORA \cite{luo2025aurora}. It is a heavyweight ensemble integrating Qwen-Omni \cite{xu2025qwen3}, VideoLLaMA2 \cite{cheng2024videollama}, and SAM. While achieving high performance through a three-stage training procedure and massive scale, it relies on complex external components and auxiliary datasets.
\end{itemize}

\subsubsection{More implementation details}
Due to space constraints in the main text, additional implementation details are provided here.
% Input video frames were resized to $224 \times 224$ before being input to MLLM.
% Differently, 
High-resolution frames were passed to SAM2 and augmented as described in \cite{ravi2024sam2}, resized to $1024 \times 1024$.
To explicitly model the target shift and maintain temporal identity, we constrained the MLLM to respond with a structured dual-token sequence: \textit{It is} \texttt{[query]} \texttt{[seg]}.
% Through end-to-end fine-tuning, the MLLM learns to decouple the target's information into two distinct latent representations.
By placing $\texttt{[query]}$ before $\texttt{[seg]}$ in the causal chain of the LLM, the model ensures that the segmentation process is conditioned on a pre-established semantic identity.
This design allows SAIL to effectively calculate and rectify the target shift within the latent space prior to mask decoding.
For each time step $t$, we encoded the refined prompt and current video frame to SAM2.
The resulting spatio-temporal features were processed by a memory encoder and stored in a memory bank as persistent anchors for future frames.
We adopted the memory attention from SAM2.
% , where the model performs cross-attention between the decoded features and the historical embeddings retrieved from the memory bank.
The final input to the mask decoder was a fusion of the current image features, the memory-attended context, and the refined embeddings.
This ensures that the generated masks are not only spatially precise but also temporally consistent with the target.

We trained SAIL in an end-to-end manner using a multi-task loss function that optimizes both the linguistic reasoning of the MLLM and the pixel-level prediction of the mask decoder.
Following the standard training protocol for MLLM-based segmentation (\textit{e.g.}, LISA \cite{lai2024lisa}), we employed an auto-regressive cross-entropy loss, $\mathcal{L}_{txt}$, for text generation.
For the binary mask prediction, we calculated the discrepancy between the predicted masks $M_t$ and the ground truth masks $\hat{M}_t$ across all $T$ frames.
The segmentation loss $\mathcal{L}_{seg}$ consists of a binary cross-entropy (BCE) loss and a DICE loss.
The final objective function $\mathcal{L}$ is defined as:
\begin{equation}
\mathcal{L} = \lambda_{txt} \mathcal{L}_{txt} + \lambda_{bce} \mathcal{L}_{bce} + \lambda_{dice} \mathcal{L}_{dice},
\end{equation}
where $\lambda_{txt}$, $\lambda_{bce}$, and $\lambda_{dice}$ are hyper-parameters.
% All experiments were conducted using PyTorch on two NVIDIA H100 GPUs.
The model was trained for 10 epochs with a batch size of 2 per device.
The hyper-parameters $\lambda_{txt}$, $\lambda_{bce}$, $\lambda_{dice}$ and $\alpha$ were set to 1.0, 1.0, 0.5 and 0.1, respectively.

\begin{table}[t]\scriptsize{}
  \caption{Comparison with existing methods on the selective samples for target shift from Ref-AVSBench \cite{wang2024refavs}.
  % We re-implemented existing methods to obtain the corresponding results (in $\ast$).
  % Note that TGS-Agent employed external Grounding-DINO \cite{liu2024grounding}.
  % $\mathcal{J}$\&$\mathcal{F}$ represents the average of $\mathcal{J}$ and $\mathcal{F}$.
  }
  \centering
  \setlength\tabcolsep{3pt}
  \begin{tabular}{p{2.5cm}|ccc|ccc|ccc}
  \toprule
  \multirow{2}*{Method} & \multicolumn{3}{c|}{Seen$\uparrow$} & \multicolumn{3}{c|}{Unseen$\uparrow$} & \multicolumn{3}{c}{Mix$\uparrow$}\\
  % \cmidrule(lr){4-5} \cmidrule(lr){6-7} \cmidrule(lr){8-9}
  
  & \textbf{$\mathcal{J}$} & \textbf{$\mathcal{F}$} & \textbf{$\mathcal{J}$}\&\textbf{$\mathcal{F}$} & \textbf{$\mathcal{J}$} & \textbf{$\mathcal{F}$}  & \textbf{$\mathcal{J}$}\&\textbf{$\mathcal{F}$} & \textbf{$\mathcal{J}$} & \textbf{$\mathcal{F}$}  & \textbf{$\mathcal{J}$}\&\textbf{$\mathcal{F}$}\\
 \midrule
  EEMC \cite{wang2024refavs}  & 17.4 & 30.5 & 24.0 & 34.2 & 52.6 & 43.4 &  25.8 & 41.6 & 33.7\\
  Crab \cite{du2025crab}  & 25.1 & 40.7 & 32.9 & 32.3 & 50.5 & 41.4 & 28.7 & 45.6 & 37.2\\
  TGS-Agent \cite{zhou2025tgs}  & 37.5 & 47.9 & 42.7 & 56.7 & 64.2 & 60.5 & 47.1 & 56.1 & 51.6\\
  AURORA \cite{luo2025aurora}  & 40.7 & 50.0 & 45.4 & 52.6 & 61.9 & 57.3 & 46.7  & 56.0 & 51.4\\
  \hline
  SAIL (ours) & \textbf{53.1} & \textbf{62.4} & \textbf{57.8} & \textbf{65.7} & \textbf{72.8} & \textbf{69.3} & \textbf{59.4} & \textbf{67.6} & \textbf{63.5}\\
  \bottomrule
  \end{tabular}
  \label{tab: comparison with SOTA for ts-ref}
  \vspace{-3mm}
\end{table}

\subsection{Target shift analysis via specialized evaluation subset}
\label{sec: additional evaluation}
To validate the effectiveness of our method in addressing the target shift problem, we curated a specialized evaluation subset derived from the Ref-AVSBench \cite{wang2024refavs} test split.
This subset focuses on challenging scenarios where the referred object undergoes transient changes in the visual or auditory domains, testing the model's ability to maintain stable consistency.
% Due to space constraints, a detailed description of this subset and an extended comparative discussion are provided in the supplementary materials.
This evaluation subset will be made \textbf{publicly available}.

\noindent
\textbf{Evaluation set.}
The selection process consists of two primary stages: automated visual filtering and manual auditory verification.
We observed that in Ref-AVSBench, the target object is consistently present and annotated in the initial frame ($t=0$).
We leveraged this as a reference point to detect potential occlusions or identity drifts.
Instead of a direct IoU comparison against the first frame, which can be misleading due to camera motion or natural target movement, we calculated the relative shrinkage ratio ($R_s$) between adjacent frames.
When $R_{s}$ falls below a predefined threshold (\textit{e.g.}, 0.3), it signified a sudden disappearance or significant reduction of the target mask, indicating a potential occlusion or a distractor-induced shift.
While the first step identifies visual disruptions, target shifts are often triggered by auditory complexity or scene density.
We manually inspected the test split to include samples that might have been missed by the geometric filter, specifically targeting scenarios with intermittent sound changes, background sounding objects, or multiple simultaneous sound sources (\textit{e.g.}, two people talking or overlapping instruments).
We also checked videos containing more than one salient object of the same category, as these possess a higher probability of identity confusion during temporal propagation.
Following this two-step pipeline, we identified 371 video sequences that exhibit significant visual or auditory shifts.
This target shift evaluation set provides a focused set to verify the effectiveness of our proposed method on target shift problem in Ref-AVS.
% To facilitate future research on temporal consistency in audio--visual tasks, this evaluation subset will be made {publicly available}.

\noindent
\textbf{Corresponding results.}
The results in Table~\ref{tab: comparison with SOTA for ts-ref} reveal a significant performance drop for all baseline methods compared to their results on the full Ref-AVSBench in Table~\ref{tab: comparison with SOTA for ref-avs}.
For instance, EEMC struggled severely on the Seen split (\textit{e.g.}, 17.4\% in $\mathcal{J}$), likely because its perception-heavy approach cannot recover once a visual target is temporarily lost.
MLLM-based methods like TGS-Agent and AURORA performed better due to their stronger reasoning backbones, yet they still exhibited substantial drift.
TGS-Agent achieved 47.1\% in $\mathcal{J}$ on the Mix split, but failed to maintain the high performance (61.3\%) for the complete test set.
% This indicates that while agents and ensembles can reason about a scene, they lack a dedicated mechanism to anchor identity in the latent space during temporal transitions.
In contrast, SAIL demonstrated a significant performance advantage, outperforming AURORA by 12.7\% in $\mathcal{J}$ on the Mix split.
Specifically, on the Unseen split, where the model cannot rely on category-specific priors, SAIL maintained a high $\mathcal{F}$ of 72.8\%.
This proves that the $\texttt{[query]}$ token effectively acts as a universal semantic anchor that identifies the target even when visual features become noisy or novel.
The consistent lead of SAIL across both Seen and Unseen scenarios in this specialized subset provides empirical evidence that our dual-token strategy and shift-aware latent refinement are not merely marginal improvements.
Instead, they represent a fundamental architectural solution to the target shift problem of identity consistency in tri-modal audio--visual scene understanding.

% \noindent
% \textbf{Visualizations.}

\subsection{More ablation studies}
\label{sec: additional abl}
% In this section, we present additional ablation studies on the fusion coefficient and mask decoder.
\subsubsection{Alternative strategies for target shift}
\label{sec: ab on alternative}
In this section, we provide detailed descriptions of the alternative strategies discussed in the main text (Sec.~\ref{sec: ablation}).
They represent three different ways: Pre-, In-, and Post-LLM processing, where target shift mitigation could potentially be implemented.
The corresponding results are reported in Table~\ref{tab: ablation on shift-aware}.
\begin{itemize}[leftmargin=*, itemsep=0pt]
\item \textit{Pre-LLM}.
This strategy focuses on enriching the input space by providing the LLM with ``object-centric'' knowledge before it begins processing.
We employed a pre-trained object detector (\textit{e.g.}, Mask R-CNN \cite{he2017maskrcnn}) to identify objects in the initial frame.
The $Top-k$ regions of interest (RoIs) were cropped and passed through a CLIP-ViT-L/14 encoder to obtain feature vectors.
These vectors were projected to the LLM's hidden dimension via a two-layer MLP and prepended as a ``visual'' token to the prompt.
As verified by the corresponding results, these explicit object embeddings fail to account for the temporal evolution of the target's appearance and sound.

\item \textit{In-LLM}.
This strategy attempts to leverage the LLM's internal self-attention to manage shifts.
We expanded the LLM with a specialized \texttt{[shift]} token, while the hidden state was passed through a linear layer to predict an estimated shift.
We then attempted a simple linear interpolation: $\mathcal{H}_{t} = \textrm{MLP}(\texttt{[seg]}_t + s\texttt{[shift]}_t)$, where $s$ is a hyper-parameter.
However, the LLM's attention mechanism is optimized for linguistic tokens, not for the high-dimensional latent shifts.
It often collapses to a mean value, failing to provide the fine-grained, channel-wise gating required to correspond specific features.

\item \textit{Post-LLM}.
This strategy explores whether the refinement should happen closer to the mask decoder than the reasoning LLM.
After the LLM, we employed a simple transformer-based module that uses cross-attention to allow $\texttt{[seg]}_t$ to search for a learnable query token from video features.
Without the global identity anchor guiding the attention, the module becomes overly sensitive to visual distractors.
In Ref-AVS scenarios, this visual-heavy refinement often locks onto the wrong sounding object.
\end{itemize}

Differently, by performing refinement in the latent space using an identity-guided anchor and shift-aware gating, our proposed SAIL maintains the semantic consistency of the target while learning multimodal changes over a video sequence.

\begin{table}[t]\scriptsize{}
  \caption{Ablation study on the shift-aware refinement module.
  }
  \centering
  \setlength\tabcolsep{3pt}
  \begin{tabular}{ccc|ccc|ccc|ccc|c}
\toprule
\multirow{2}{*}{Fusion} & \multirow{2}{*}{Residual} & \multirow{2}{*}{Gating weights} & \multicolumn{3}{|c}{Seen$\uparrow$} & \multicolumn{3}{|c}{Unseen$\uparrow$} & \multicolumn{3}{|c|}{Mix$\uparrow$} & \multicolumn{1}{c}{Null$\downarrow$}\\
 & & & \textbf{$\mathcal{J}$} & \textbf{$\mathcal{F}$} & \textbf{$\mathcal{J}$}\&\textbf{$\mathcal{F}$} & \textbf{$\mathcal{J}$} & \textbf{$\mathcal{F}$} & \textbf{$\mathcal{J}$}\&\textbf{$\mathcal{F}$} & \textbf{$\mathcal{J}$} & \textbf{$\mathcal{F}$} & \textbf{$\mathcal{J}$}\&\textbf{$\mathcal{F}$} & \textbf{$\mathcal{S}$}\\
 \midrule
 \checkmark & & & 60.9 & 70.5 & 65.7 & 65.4 & 72.7 & 69.1 & 63.2 & 71.6 & 67.4 & 0.027\\
 & \checkmark &  & 66.3 & 75.4 & 70.9 & 70.3 & 77.8 & 74.1 & 68.3 & 76.6 & 72.5 & 0.015\\
  & \checkmark & \checkmark & \textbf{68.4} & \textbf{77.5} & \textbf{73.0} & \textbf{72.0} & \textbf{79.5} & \textbf{75.8} & \textbf{70.2} & \textbf{78.5} & \textbf{74.4} & \textbf{0.010}\\
\bottomrule
\end{tabular}
  \label{tab: ablation on shift-aware refinment}
  \vspace{-3mm}
\end{table}

\subsubsection{Ablation on shift-aware refinement module}
\label{sec: ab on shift-aware refinment}
To verify the effectiveness of the proposed residual-based refinement and the dynamic gating mechanism, we conducted a structural ablation study.
We compared our design against a simple fusion variance, where the segmentation token is updated via concatenation with modality evidence after projection, rather than calculating the identity drift.
As shown in Table~\ref{tab: ablation on shift-aware refinment}, replacing our residual-based drift estimation (\textit{i.e.}, $\Delta$) with a simple fusion strategy led to a significant performance drop (\textit{e.g.}, $\mathcal{J}\&\mathcal{F}$ from 72.5\% to 67.4\% on the Mix split).
This confirms the intuition that in the presence of target shift, simply ``adding'' modality features is insufficient.
The residual design acts as a targeted error-correction signal, specifically identifying and rectifying the displacement between the current perceived evidence and the global identity anchor.
The inclusion of modality-specific gating weights (\textit{i.e.}, $\lambda$) further enhanced the model's robustness, improving $\mathcal{J}\&\mathcal{F}$ by $1.9\%$ on the Mix split.
This demonstrates that the dynamic weights effectively suppress noisy residuals from unreliable modalities, such as audio segments with high background interference or visual frames with heavy occlusion, ensuring that only high-confidence evidence is used to refine the latent states.
Overall, the combination of residual-based drift estimation and gating provided the most stable performance across both Seen and Unseen splits.
This architecture ensures that SAIL not only captures the semantic identity but also accurately maintains the spatial localization even when the tri-modal signals are temporarily inconsistent.

\begin{table}[t]\scriptsize{}
  \caption{Ablation study on LMMs.
  We replaced the MLLM in our SAIL with LLaMA2 \cite{touvron2023llama} and InternVL2 \cite{chen2024internvl2}, respectively.
  }
  \centering
  \setlength\tabcolsep{4pt}
  \begin{tabular}{p{2.3cm}|ccc|ccc|ccc|c}
\toprule
\multirow{2}{*}{Method} & \multicolumn{3}{|c}{Seen$\uparrow$} & \multicolumn{3}{|c}{Unseen$\uparrow$} & \multicolumn{3}{|c|}{Mix$\uparrow$} & \multicolumn{1}{c}{Null$\downarrow$}\\
 & \textbf{$\mathcal{J}$} & \textbf{$\mathcal{F}$} & \textbf{$\mathcal{J}$}\&\textbf{$\mathcal{F}$} & \textbf{$\mathcal{J}$} & \textbf{$\mathcal{F}$} & \textbf{$\mathcal{J}$}\&\textbf{$\mathcal{F}$} & \textbf{$\mathcal{J}$} & \textbf{$\mathcal{F}$} & \textbf{$\mathcal{J}$}\&\textbf{$\mathcal{F}$} & \textbf{$\mathcal{S}$}\\
 \midrule
 LLaMA2-7B \cite{touvron2023llama} & 65.1 & 73.7 & 69.4 & 69.2 & 76.7 & 73.0 & 67.2 & 75.2 & 71.2 & 0.028\\
 InternVL2-8B \cite{chen2024internvl2} & 67.6 & 76.8 & 72.2 & 70.6 & 77.4 & 74.0 & 69.1 & 77.1 & 73.1 & 0.014\\
 Chat-UniVi-7B \cite{jin2024chatunivi} & \textbf{68.4} & \textbf{77.5} & \textbf{73.0} & \textbf{72.0} & \textbf{79.5}  & \textbf{75.8} & \textbf{70.2} & \textbf{78.5} & \textbf{74.4} & \textbf{0.010}\\
\bottomrule
\end{tabular}
  \label{tab: ablation on llms}
  \vspace{-3mm}
\end{table}

\subsubsection{Ablation on LLMs}
\label{sec: ab on llm}
To verify that the performance gains of SAIL are rooted in our proposed shift-aware refinement rather than the specific choice of the LLM backbone, we ablated the MLLM component.
We replaced the Chat-UniVi \cite{jin2024chatunivi} with LLaMA2 \cite{touvron2023llama} (a text-centric LLM) and InternVL2 \cite{chen2024internvl2} (a high-capacity vision-language model).
The results in Table~\ref{tab: ablation on llms} indicate that SAIL maintained strong performance regardless of the underlying MLLMs.
Even with LLaMA2, the model achieved a $\mathcal{J}$ of 67.2\% on the Mix split, compared to Crab (43.1\%) that also used the LLM.
% This suggests that our identity-guided anchoring and latent refinement modules are capable of providing the necessary spatial-temporal grounding even when the linguistic prior is not natively aligned with video features.
The model with Chat-UniVi yielded the best results across all metrics.
% We attribute this to several key factors: (1) 
Unlike InternVL2, which is optimized for static images, Chat-UniVi is specifically designed for unified visual representation across images and videos.
Its ability to handle temporal sequences more fluidly provides a more stable initial tokens for our refinement module.
% While InternVL2 is a significantly larger model, it may introduce more noise on the Null split.
% This suggests that for pixel-wise segmentation tasks, the balance between high-level reasoning and fine-grained feature preservation in Chat-UniVi is more compatible with the SAIL framework.
LLaMA2's lower performance on the Null split highlights the disadvantage of using a model that lacks integrated multimodal pre-training.
% The higher "Null" split error indicates that without a strong multimodal foundation, the model is more prone to hallucinating segments when no sounding object is present.
In summary, while a stronger MLLM backbone can provide a better starting point, the consistent success of SAIL across different architectures confirms that our dual-token latent refinement is the primary driver of temporal consistency and target shift rectification.

\subsubsection{Ablation on fusion coefficient}
\label{sec: ab on fusion co}
We investigated the sensitivity of the SAIL framework to the fusion coefficient $\alpha$ on Ref-AVSBench \cite{wang2024refavs}, which governs the intensity of the identity-guided refinement applied to the latent segmentation tokens.
As shown in Table~\ref{tab: ablation on fusion coeff}, the model achieved peak performance across all metrics when $\alpha$ is set to $0.1$.
The results reveal a consistent performance degradation as the value of $\alpha$ increases.
Specifically, increasing $\alpha$ from $0.1$ to $1.0$ led to a notable drop in $\mathcal{J}$, from 70.2\% to 66.0\% on the Mix split.
This trend suggests that while anchor is crucial for target identity, an excessively large coefficient may over-correct the latent embeddings, potentially overriding the spatial cues captured by the original segmentation token.
Furthermore, the Null split performance was most stable at $\alpha=0.1$, indicating that a carefully calibrated refinement better preserves the model's ability to suppress false-positive segments when no valid target is present.
Consequently, we adopt $\alpha=0.1$ as the default setting to maintain an optimal balance between identity anchoring and segmentation representation.

\begin{table}[t]\scriptsize{}
  \caption{Ablation study on the fusion coefficient $\alpha$ in the anchor alignment before the mask decoder.
  }
  \centering
  \setlength\tabcolsep{3pt}
  \begin{tabular}{p{1.5cm}|ccc|ccc|ccc|c}
\toprule
\multirow{2}{*}{Method} & \multicolumn{3}{|c}{Seen$\uparrow$} & \multicolumn{3}{|c}{Unseen$\uparrow$} & \multicolumn{3}{|c|}{Mix$\uparrow$} & \multicolumn{1}{c}{Null$\downarrow$}\\
 & \textbf{$\mathcal{J}$} & \textbf{$\mathcal{F}$} & \textbf{$\mathcal{J}$}\&\textbf{$\mathcal{F}$} & \textbf{$\mathcal{J}$} & \textbf{$\mathcal{F}$} & \textbf{$\mathcal{J}$}\&\textbf{$\mathcal{F}$} & \textbf{$\mathcal{J}$} & \textbf{$\mathcal{F}$} & \textbf{$\mathcal{J}$}\&\textbf{$\mathcal{F}$} & \textbf{$\mathcal{S}$}\\
 \midrule
 $\alpha=0.1$ & \textbf{68.4} & \textbf{77.5} & \textbf{73.0} & \textbf{72.0} & \textbf{79.5}  & \textbf{75.8} & \textbf{70.2} & \textbf{78.5} & \textbf{74.4} & \textbf{0.010}\\
 $\alpha=0.2$ & 67.2 & 75.0 & 71.1 & 70.8 & 78.2 & 74.5 & 69.0 & 76.6 & 72.8 & 0.011\\
 $\alpha=0.5$ & 66.7 & 74.3 & 70.5 & 70.1 & 77.6 & 73.9 & 68.4 & 76.0 & 72.2 & 0.012\\
 $\alpha=1.0$ & 63.4 & 72.4 & 67.9 & 68.5 & 76.7 & 72.6 & 66.0 & 74.6 & 70.3 & 0.017\\
\bottomrule
\end{tabular}
  \label{tab: ablation on fusion coeff}
  \vspace{-3mm}
\end{table}

\begin{table}[t]\scriptsize{}
  \caption{Ablation study on the mask decoder.
  QP denotes the query propagation operation in OISA-1B \cite{ying2025oisa-1b}.
  }
  \centering
  \setlength\tabcolsep{3pt}
  \begin{tabular}{p{1.5cm}|ccc|ccc|ccc|c}
\toprule
\multirow{2}{*}{Method} & \multicolumn{3}{|c}{Seen$\uparrow$} & \multicolumn{3}{|c}{Unseen$\uparrow$} & \multicolumn{3}{|c|}{Mix$\uparrow$} & \multicolumn{1}{c}{Null$\downarrow$}\\
 & \textbf{$\mathcal{J}$} & \textbf{$\mathcal{F}$} & \textbf{$\mathcal{J}$}\&\textbf{$\mathcal{F}$} & \textbf{$\mathcal{J}$} & \textbf{$\mathcal{F}$} & \textbf{$\mathcal{J}$}\&\textbf{$\mathcal{F}$} & \textbf{$\mathcal{J}$} & \textbf{$\mathcal{F}$} & \textbf{$\mathcal{J}$}\&\textbf{$\mathcal{F}$} & \textbf{$\mathcal{S}$}\\
 \midrule
 + M2F & 62.8 & 71.9 & 67.4 & 65.8 & 73.2 & 69.5 & 64.3 & 72.6 & 68.5 & 0.020\\
 + SAM & 64.3 & 73.8 & 69.1 & 67.7 & 75.0 & 71.4 & 66.0 & 74.4 & 70.2 & 0.015 \\
 + QP + SAM & 66.9 & 76.1 & 71.5 & 71.0 & 78.5 & 74.8 & 69.0 & 77.3 & 73.2 & \textbf{0.010}\\
 + SAM2 & 65.0 & 75.2 & 70.1 & 70.4 & 78.0 & 74.2 & 67.7 & 76.6 & 72.2 & 0.013 \\
 % \hline
 SAIL(ours) & \textbf{68.4} & \textbf{77.5} & \textbf{73.0} & \textbf{72.0} & \textbf{79.5} & \textbf{75.8} & \textbf{70.2} & \textbf{78.5} & \textbf{74.4} & \textbf{0.010}\\
\bottomrule
\end{tabular}
  \label{tab: ablation on mask decoder}
  \vspace{-3mm}
\end{table}

\subsubsection{Ablation on mask decoder}
\label{sec: ab on decoder}
We further evaluated the impact of different mask decoders on the segmentation performance.
Specifically, we replaced the specialized mask decoder in our SAIL with four baseline configurations: a standard Mask2Former (M2F) decoder \cite{cheng2022mask2former}, a vanilla Segment Anything Model (SAM) decoder \cite{kirillov2023sam}, a query propagation (QP) enhanced SAM decoder as proposed in OISA-1B, and a SAM2 decoder \cite{ravi2024sam2} as adopted in SAM2-LOVE.
As presented in Table~\ref{tab: ablation on mask decoder}, the transition from M2F to SAM yielded a noticeable improvement, demonstrating the superior zero-shot localization capabilities of the foundation decoder.
The inclusion of QP, which enables frame-by-frame online updates of the query to capture temporal motion and contextual information, further boosted the performance.
This highlights the importance of modeling the target's dynamic evolution across the video timeline.
Similarly, the inclusion of SAM2 also achieved higher performance than static M2F or SAM, especially on the Unseen split.
Notably, our proposed SAIL outperformed all the above configurations, achieving the highest scores across all metrics.
While QP focuses on smooth motion tracking through query updates and SAM2 employs memory bank updates, SAIL's specialized mask decoder enables shift-aware prompting and memory-based propagation that provide a more robust mechanism for consistent segmentation, leading to 1.2\% and 2.2\% improvements in $\mathcal{J}\&\mathcal{F}$ on the Mix split, respectively.
The results confirm that SAIL's specialized mask decoder effectively integrates shift-aware representations, enabling more robust and consistent segmentation in complex multimodal scenarios.

\begin{figure}
    \centering
    \begin{subfigure}[t]{0.95\linewidth}
    \includegraphics[width=1\linewidth]{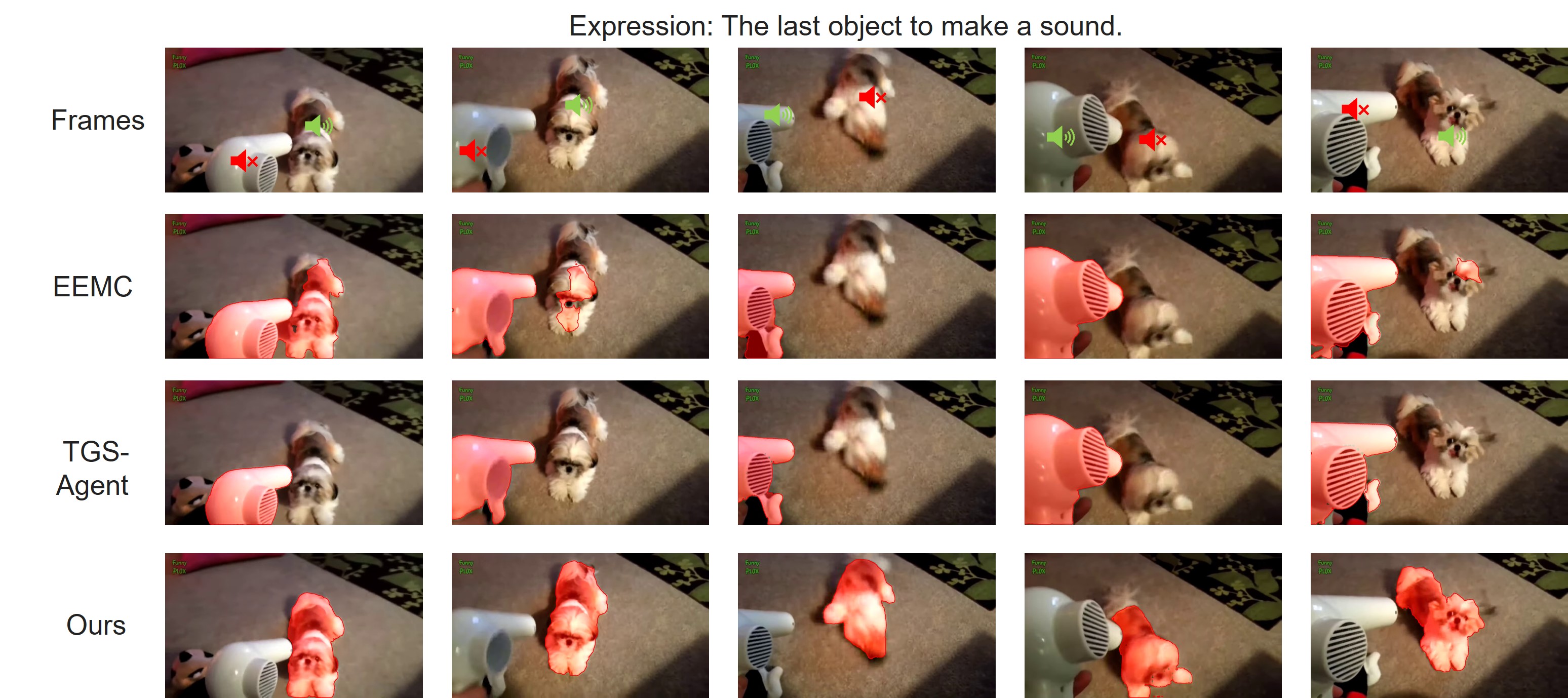}
    \subcaption{}
  \end{subfigure}
  % \hspace{5mm}
  \begin{subfigure}[t]{0.95\linewidth}
    \includegraphics[width=1\linewidth]{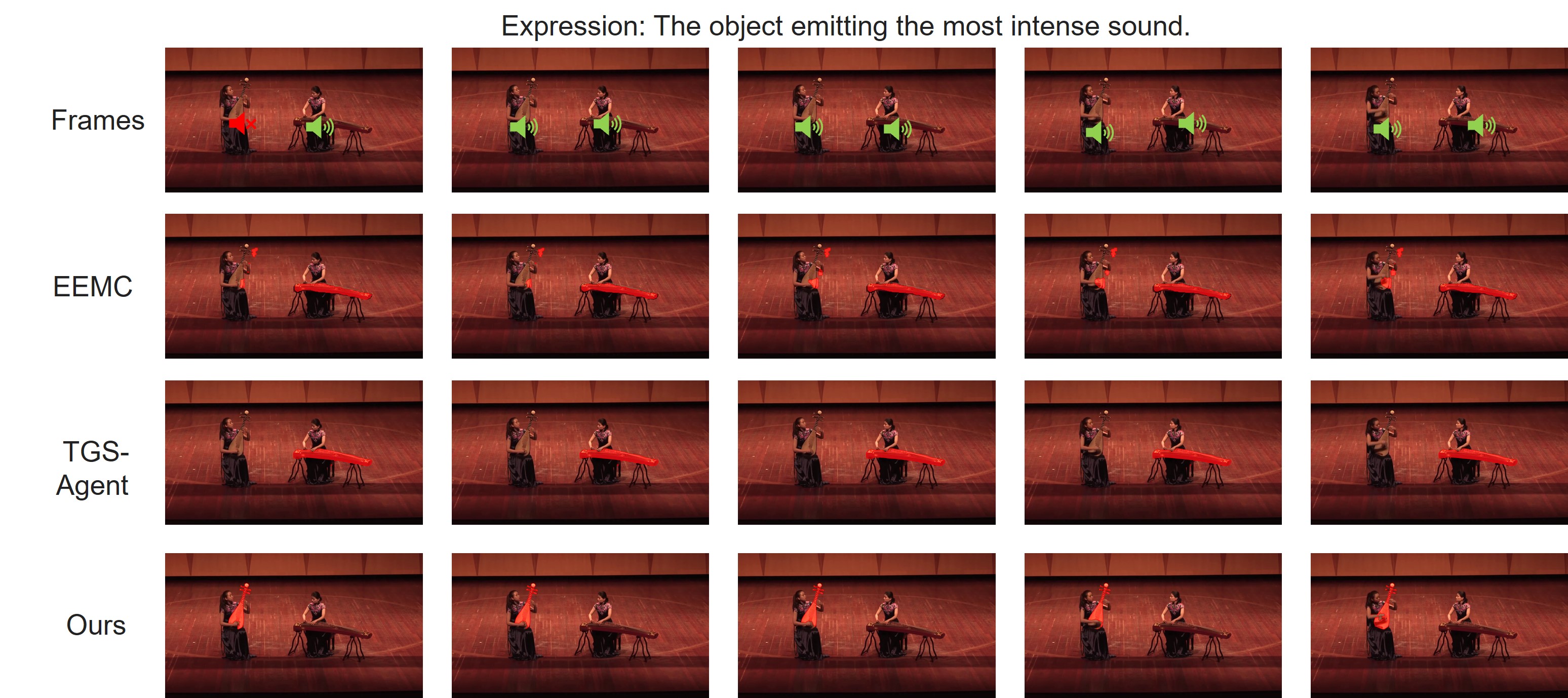}
    \subcaption{}
  \end{subfigure}
    \caption{Additional qualitative results from EEMC \cite{wang2024refavs}, TGS-Agent \cite{zhou2025tgs}, and our proposed SAIL on the Seen split from the Ref-AVSBench \cite{wang2024refavs} dataset.
    }
    \vspace{-5mm}
    \label{fig: more vis seen}
\end{figure}

\begin{figure}
    \centering
    \begin{subfigure}[t]{0.95\linewidth}
    \includegraphics[width=1\linewidth]{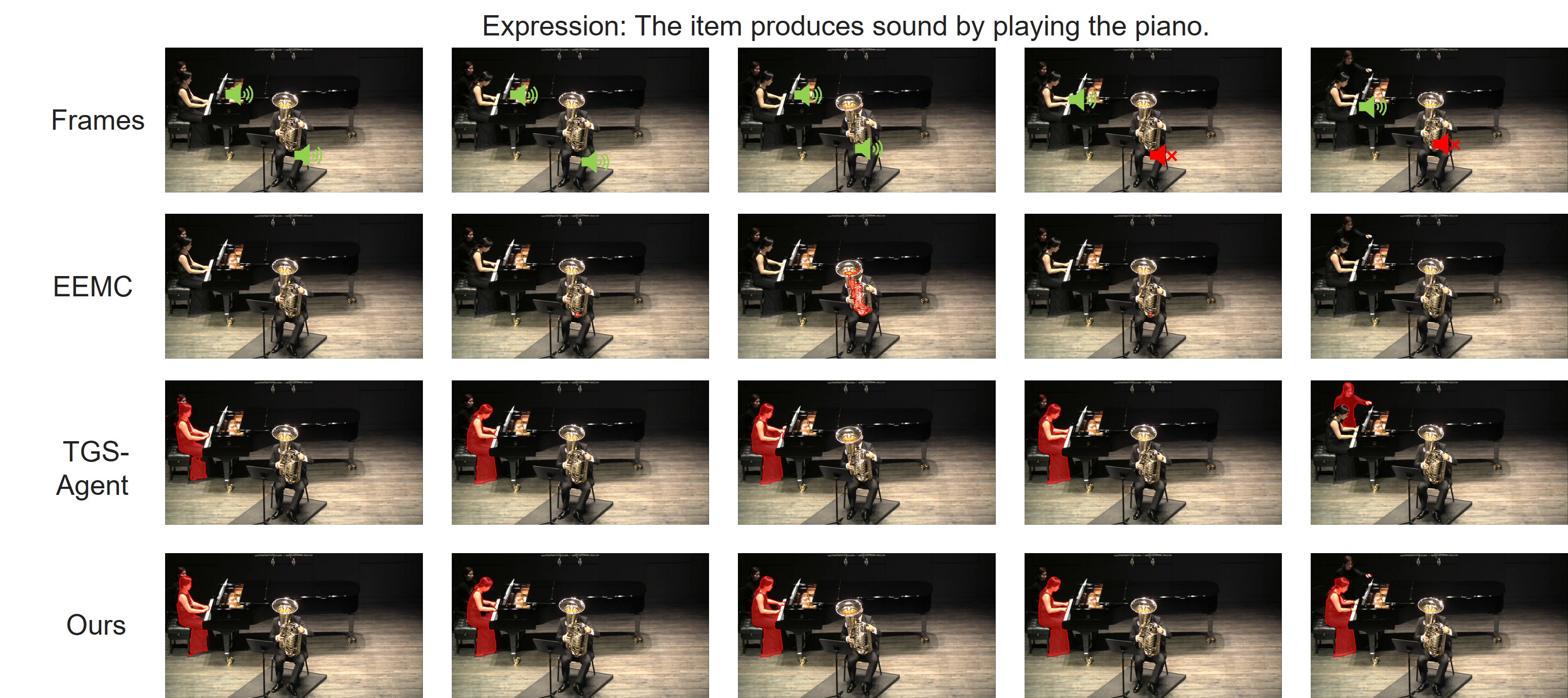}
    \subcaption{}
  \end{subfigure}
  % \hspace{5mm}
  \begin{subfigure}[t]{0.95\linewidth}
    \includegraphics[width=1\linewidth]{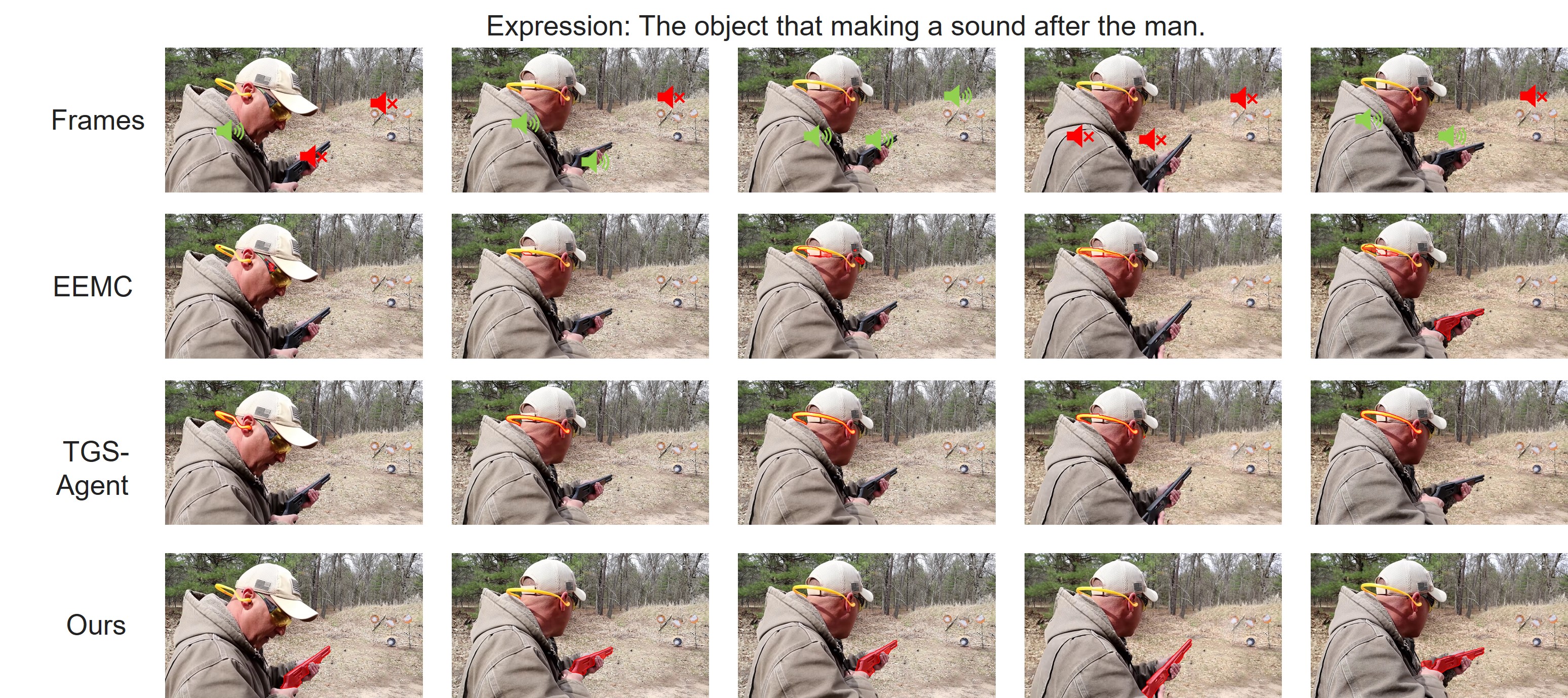}
    \subcaption{}
  \end{subfigure}
    \caption{Additional qualitative results from EEMC \cite{wang2024refavs}, TGS-Agent \cite{zhou2025tgs}, and our proposed SAIL on the Unseen split from the Ref-AVSBench \cite{wang2024refavs} dataset.
    }
    \vspace{-5mm}
    \label{fig: more vis unseen}
\end{figure}

\begin{figure}
    \centering
    \begin{subfigure}[t]{0.95\linewidth}
    \includegraphics[width=1\linewidth]{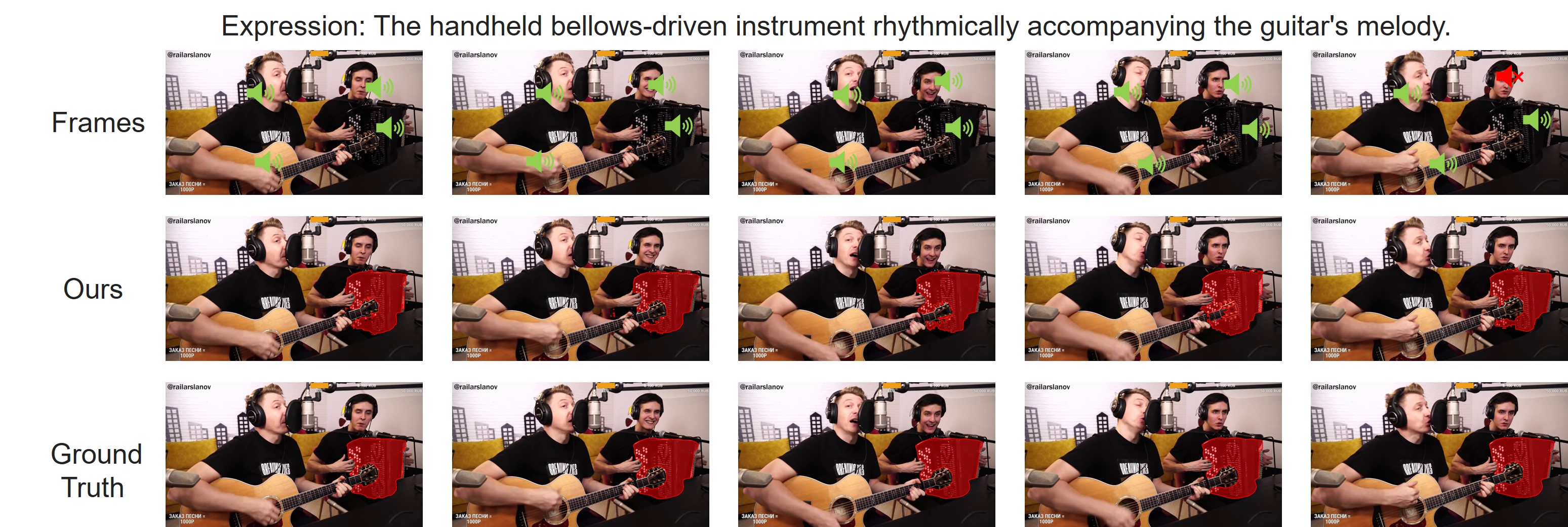}
    \subcaption{}
  \end{subfigure}
  % \hspace{5mm}
  \begin{subfigure}[t]{0.95\linewidth}
    \includegraphics[width=1\linewidth]{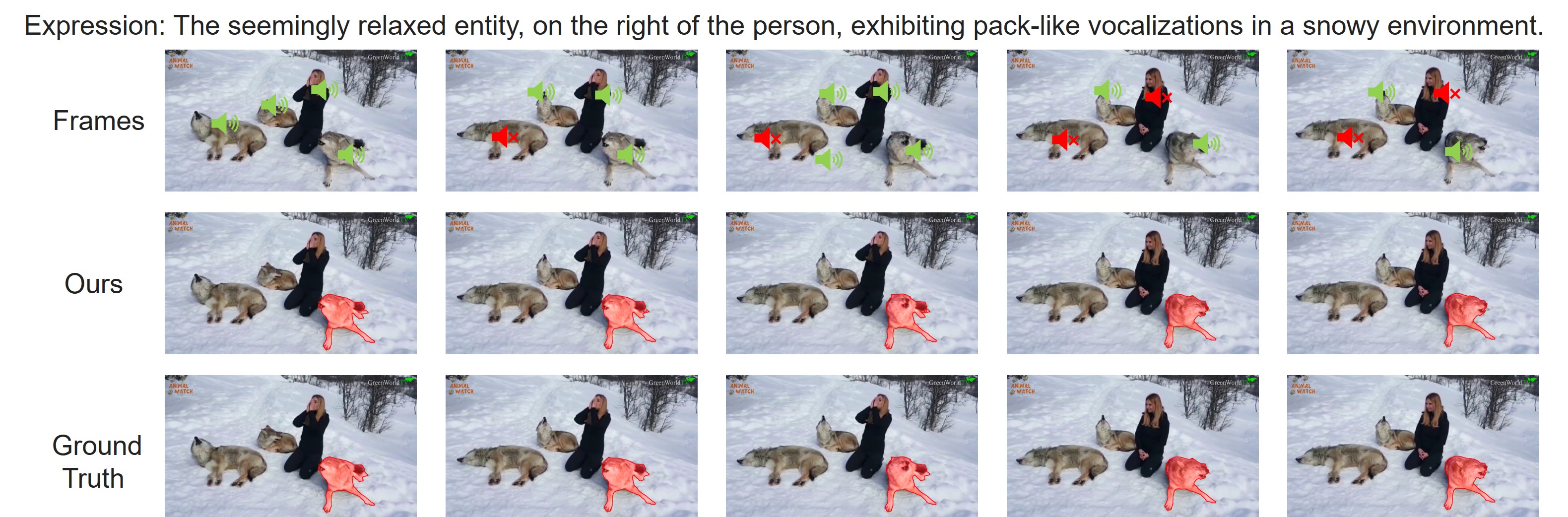}
    \subcaption{}
  \end{subfigure}
    \caption{Qualitative results on samples from the $\mathrm{R^2}$-AVSBench \cite{zhou2025tgs} dataset.
    The two examples are from the Seen and Unseen splits, respectively.
    }
    \vspace{-5mm}
    \label{fig: r2}
\end{figure}

\begin{figure}
    \centering
    \begin{subfigure}[t]{0.95\linewidth}
    \includegraphics[width=1\linewidth]{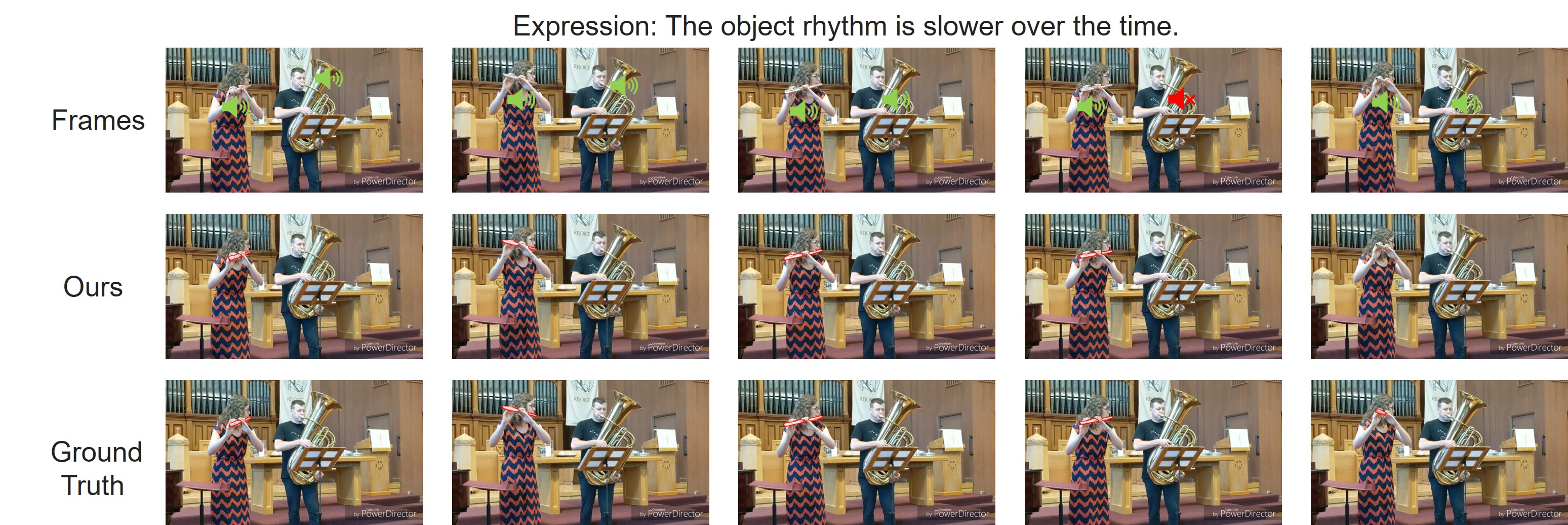}
    \subcaption{}
  \end{subfigure}
  % \hspace{5mm}
  \begin{subfigure}[t]{0.95\linewidth}
    \includegraphics[width=1\linewidth]{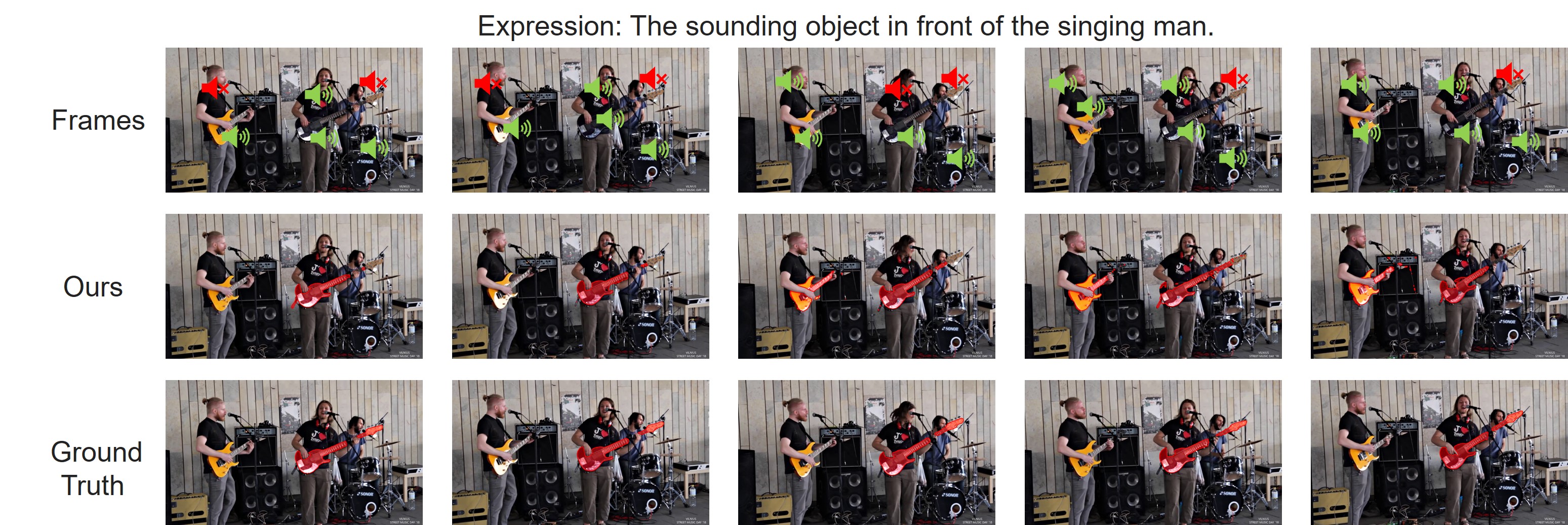}
    \subcaption{}
  \end{subfigure}
    \caption{Visualization of failure cases for the proposed SAIL on the Ref-AVSBench \cite{wang2024refavs} dataset.
    }
    \vspace{-5mm}
    \label{fig: failures}
\end{figure}

\subsection{More qualitative results}
\label{sec: more vis}
In this section, we present more qualitative results in Fig.~\ref{fig: more vis seen} and Fig.~\ref{fig: more vis unseen}, illustrating the model's performance on the Seen and Unseen splits of Ref-AVSBench \cite{wang2024refavs}, respectively.
For each example, we show video frames, expression, and predictions by different models.
The visualized examples highlight SAIL's superior capability in reasoning over complex tri-modal inputs while maintaining temporal identity consistency.
In Fig.~\ref{fig: more vis seen}(a), the scene contains two potential targets.
While EEMC failed to localize the focus of the expression, which was distracted by the visual prominence of the hair-dryer, and TGS-Agent mistakenly identified the hair-dryer as the primary sounding object, SAIL correctly segmented the dog.
This success is attributed to our shift-aware refinement, which leverages the \texttt{[query]} token to filter out acoustic distractors and maintain focus on the linguistically referred target.
Fig.~\ref{fig: more vis unseen}(b) depicts a challenging out-of-distribution scenario involving a ``gun'' and an ``earphone''.
Considering the presence of a salient yellow earphone and the explicit phrasing ``after the man'', TGS-Agent was misled by the visual distractor of the earphone.
In contrast, SAIL generated consistent and accurate masks for the target throughout the timeline.
This confirms that our identity-guided anchoring effectively generalizes to novel categories by prioritizing semantic logic over raw visual bias.

We further evaluated SAIL on the $\mathrm{R^2}$-AVSBench \cite{zhou2025tgs} to qualitatively assess its performance in scenarios requiring complex semantic reasoning.
Fig.~\ref{fig: r2} presents two examples for implicit expressions.
Even when the target is described through indirect functional or acoustic attributes, our method demonstrated remarkable robustness in preserving target identity.
This visual evidence confirms that the shift-aware refinement mechanism effectively bridges the gap between linguistic reasoning and pixel-wise consistency, maintaining a stable mask even as the visual and auditory contexts evolve.

Despite its robust performance, SAIL encounters limitations in specific edge cases, as illustrated in Fig.~\ref{fig: failures}.
As shown in the top example of the figure, when a target is both relatively small and partially occluded (\textit{e.g.}, the flute hold by the woman in the final frame), the refined segmentation tokens may lack sufficient spatial density to guide the mask decoder accurately.
In such instances, the latent information is dominated by background features, leading to fragmented or missing masks.
In scenarios involving multiple overlapping sound sources paired with imprecise or underspecified linguistic expressions (\textit{e.g.}, Fig.~\ref{fig: failures}(b)), the model may suffer from target ambiguity.
When the prompt does not sufficiently distinguish between two active sounding objects, the model may default to the most visually salient one, leading to an identity drift.
This limitation is further exacerbated in tri-modal settings, where ambiguity in referring expressions, coupled with competing auditory and visual cues, makes consistent target grounding particularly challenging.
These failure modes suggest that while our latent refinement significantly mitigates target shift, there remains room to improve the model's sensitivity to fine-grained spatial details and its ability to resolve high-level semantic contradictions.
We leave the integration of multi-scale spatial feedback and interactive disambiguation for future work.

\subsection{Complexity analysis}
\label{sec: computation}

\begin{wraptable}{r}{0.45\linewidth}\scriptsize{}
\vspace{-10pt} 
\centering
\setlength\tabcolsep{3pt}
\begin{tabular}{l|c|c|c}
\toprule
Method & Latency (ms) & FPS & Memory (GB)\\
\midrule
SAM2-LOVE & 54 & 18.5 & 12.4\\
Crab & 115 & 8.7 & 18.2\\
SAIL & 128 & 7.8 & 22.5\\
\bottomrule
\end{tabular}
\caption{Latency and memory usage of different models.}
\label{tab: complexity}
\vspace{-10pt}
\end{wraptable}
To evaluate computational efficiency, we compared SAIL with two representative baselines, SAM2-LOVE and Crab.
All measurements were conducted on a single NVIDIA H100 GPU with a batch size of 1 for fair comparison.
As shown in Table~\ref{tab: complexity}, SAIL incurs higher computational overhead than SAM2-LOVE and Crab, with an inference latency of 128 ms, throughput of 7.8 FPS, and peak memory usage of 22.5 GB.
This additional cost mainly stems from the shift-aware refinement and identity-guided reasoning modules introduced to mitigate target shift.
Compared with Crab, SAIL introduces only a moderate latency increase (+13 ms) while yielding substantial performance gains on challenging target-shift scenarios, achieving a +26.3\% improvement in $\mathcal{J}\&\mathcal{F}$ on the Mix split in Table~\ref{tab: comparison with SOTA for ts-ref}.
Notably, SAIL is trained end-to-end and performs inference without post-processing, additional instruction-tuning data, or multi-stage pipelines with external components.

\subsection{Discussion and future work}
\label{sec: discussion}
While SAIL demonstrates state-of-the-art performance in Ref-AVS, several avenues remain for further exploration and refinement.
A primary limitation of our current framework is its reliance on per-frame token generation.
While this approach ensures high precision for the 10-second clips typical of the Ref-AVSBench, it presents a computational bottleneck for long-form video analysis.
Processing extended sequences requires more efficient temporal abstraction.
Future work could draw inspiration from Ref-VOS techniques \cite{wu2022referformer, fu2025objectrelator}, implementing key-frame selection and sparse-to-dense latent propagation.
In addition, SAIL operates in a \textit{one-shot} inference mode, where the model processes a static prompt and returns a set of masks.
However, human-AI collaboration often requires a more dynamic and interactive referring process \cite{zhao2025unifiedis}.
We envision extending SAIL to support multi-turn dialogues where users can provide corrective feedback (\textit{e.g.}, ``No, the person on the left'') or adjust the segmentation focus through iterative natural language refinement.
Developing a more intelligent and controllable interface would allow SAIL to operate as a proactive assistant in complex video editing and processing contexts.
Last but not least, fine-grained multi-object relational reasoning is a promising direction for Ref-AVS.
Integrating scene-graph reasoning \cite{nguyen2024scenegraph} or object-level relational \cite{li2025post} modules could strengthen the model's fine-grained spatial understanding.

%%%%%%%%%%%%%%%%%%%%%%%%%%%%%%%%%%%%%%%%%%%%%%%%%%%%%%%%%%%%

% \newpage
% \input{checklist.tex}

\end{document}